\documentclass[preprint,12pt]{elsarticle}

\usepackage[nothing]{algorithm}
\usepackage[noend]{algpseudocode}
\usepackage{amsmath} 
\usepackage{amssymb}  
\usepackage{amsthm}
\usepackage{graphicx}
\usepackage{tabularx}
\usepackage{booktabs}
\usepackage{xcolor}
\usepackage{lineno}

\newcommand{\appname}{LiODOM}

\journal{Robotics and Autonomous Systems}

\newcommand{\change}[1]{{#1}}

\begin{document}

\begin{frontmatter}

\title{LiODOM: Adaptive Local Mapping for Robust LiDAR-Only Odometry}

\author[uib,idisba]{Emilio Garcia-Fidalgo\corref{cor1}}
\ead{emilio.garcia@uib.es}

\author[uib,idisba]{Joan P. Company-Corcoles}
\ead{joan.pep.company@gmail.com}

\author[uib,idisba]{Francisco Bonnin-Pascual}
\ead{xisco.bonnin@uib.es}

\author[uib,idisba]{Alberto Ortiz}
\ead{alberto.ortiz@uib.es}

\cortext[cor1]{Corresponding author.}

\affiliation[uib]{organization={Department of Mathematics and Computer Science, University of the Balearic Islands},
            addressline={Ctra. Valldemossa km 7.5}, 
            city={Palma de Mallorca},
            postcode={07122}, 
            country={Spain}}
            
\affiliation[idisba]{organization={Institut d'Investigacio Sanitaria Illes Balears, Hospital Universitario Son Espases},
            addressline={Ctra. Valldemossa 79}, 
            city={Palma de Mallorca},
            postcode={07120}, 
            country={Spain}}

\begin{abstract}
In the last decades, Light Detection And Ranging (LiDAR) technology has been extensively explored as a robust alternative for self-localization and mapping. These approaches typically state ego-motion estimation as a non-linear optimization problem dependent on the correspondences established between the current point cloud and a map, whatever its scope, local or global. This paper proposes \appname{}, a novel LiDAR-only ODOmetry and Mapping approach for pose estimation and map-building, based on minimizing a loss function derived from a set of weighted point-to-line correspondences with a local map abstracted from the set of available point clouds. Furthermore, this work places a particular emphasis on map representation given its relevance for quick data association. To efficiently represent the environment, we propose a data structure that combined with a hashing scheme allows for fast access to any section of the map. \appname{} is validated by means of a set of experiments on public datasets, for which it compares favourably against other solutions. Its performance on-board an aerial platform is also reported.
\end{abstract}


\end{frontmatter}


\section{Introduction}
\label{sec:intro}
Self-localization and mapping, either performed simultaneously or in a sequential fashion, are crucial abilities for a mobile robot to be useful in relevant applications, irrespective of whether the robot operates fully autonomously or in a semi-autonomous way. As stated many years ago, odometry estimation is a fundamental piece within this framework. A plethora of sensing devices have been adopted throughout the years, comprising tachometers/wheel encoders, inertial and heading sensors, time of flight sensors, and motion estimation devices, to name but a few. Among all of them, laser scanners and, for a few years now, cameras have turned out to be the sensors of choice. The latter have been extensively used~\cite{Mur2017,Ferrera2021} due to the rich perception of the surrounding world encoded in images. Vision-based estimation is however sensitive to lighting conditions, have a limited horizontal field of view and require additional calculations to acquire depth and shape perception. In contrast, 3D laser scanners provide a 360-degree overview of the platform surroundings, supply reliable range estimations and, especially motivated by the development of self-driving cars, recently have become an affordable choice for pose estimation and mapping.

LiDAR odometry is typically stated as an optimization problem that is solved using the Iterative Closed Point (ICP) algorithm~\cite{Besl1992} or any of its variants. For this to happen in a  satisfactory, fast and accurate way, a set of reliable correspondences between the current point cloud and a map must be found. A KD-tree is a popular choice to represent the whole map~\cite{Zhang14}, although the resulting performance degrades as the number of points to be managed increases, what makes necessary a filtering step to screen most relevant points. An alternative is to build a local map using a sliding window~\cite{Ye2019,Shan2020}, although this might discard useful associations that could be found if the search was performed over a global map.

\begin{figure}[tb]
\centering
\includegraphics[width=0.85\columnwidth,clip,trim=50 40 60 10]{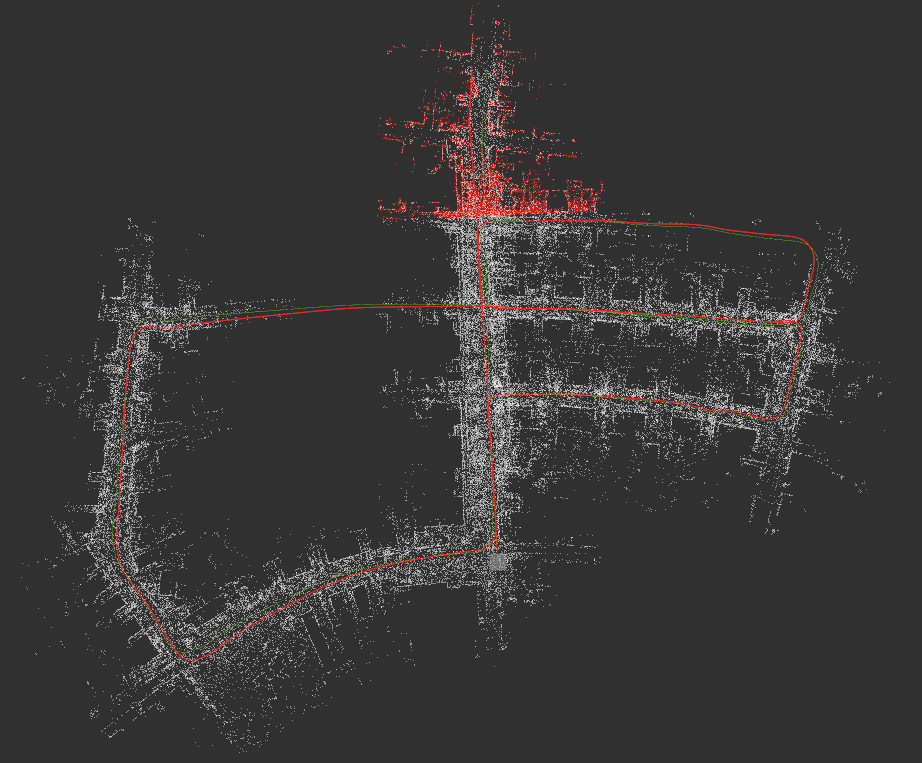}
\caption{Example of map produced by \appname{} (KITTI 05 sequence), comprising an unoptimized global map generated during navigation (in white) and a local map (in red) that is retrieved according to the position of the vehicle, to be used for next pose estimation.}
\label{fig:firstilust}
\end{figure}

This paper proposes \appname{}, a novel LiDAR odometry and mapping approach that is able to estimate the pose without additional sensors, e.g. IMU and/or GPS, unlike other recent approaches~\cite{Ye2019,Shan2020}. Our approach is formulated as a non-linear optimization problem based on a set of point-to-line constraints, weighted according to the distance from each point to the sensor center. Furthermore, we propose an efficient data structure, based on a hashing scheme, to represent the map. As a result, a local map can be retrieved according to the pose of the robot in an effective way, and point cloud correspondences against the local map can be efficiently established. This \textit{adaptive} solution naturally allows us to find correspondences between the current point cloud and revisited places (contrary to just using a sliding window). Figure~\ref{fig:firstilust} illustrates the performance of \appname{}.

In brief, the main contributions of this work are:
\begin{itemize}
    \item A LiDAR-only odometry framework that is based on an optimization problem supported by weighted point-to-line factors computed from the correspondences with a local map.
    \item A fast and efficient mapping approach based on a hash-based data structure that speeds up searches and permits to gracefully update large-scale maps.
    \item An extensive evaluation of the proposed approach on public datasets, including a comparison with other state-of-the-art methods. 
     \item An additional evaluation of \appname{} on-board an aerial platform where it provides, among others, speed estimations that are used inside velocity control loops, what shows \appname{}'s capability to operate in real time.
    \item A final contribution is the availability for the community of the source code\footnote{http://github.com/emiliofidalgo/liodom} of our approach.
\end{itemize}

The rest of the paper is organized as follows: Section~\ref{sec:relwork} overviews most relevant works in the field; the proposed framework is introduced in Sections~\ref{sec:sysoverview},~\ref{sec:odom} and~\ref{sec:mapping}; Section~\ref{sec:results} reports on the results obtained; to finish, Section~\ref{sec:conclusions} concludes the paper and suggests future research lines.

\section{Related Work}
\label{sec:relwork}

\change{Laser scanners and vision cameras are the two sensor modalities that have attracted the largest number of researchers in the last years because of their respective success stories and well-known advantages regarding motion estimation ---briefly speaking, reliable and scale-aware depth measurements in the first case, and low weight, low power and low cost in the second case. Although \appname{} belongs to the first class, in the following, we not only revise most recent laser-based works close to our solution, but also others relying on imaging sensors in order to gain perspective on the respective methodologies and achievements, without being intent to be exhaustive. Finally, we also consider map indexing.}

\subsection{\change{LiDAR-based Solutions}}

Most recent approaches carry out LiDAR-based odometry in combination with an IMU for higher accuracy. These solutions are typically regarded as loosely- and tightly-coupled methods~\cite{Shan2020,Qin2020lins}. Loosely-coupled methods estimate the state from each sensor separately. Arguably the most well-known method that falls into this category is LiDAR Odometry and Mapping (LOAM)\cite{Zhang14}, where \textit{edges} and \textit{surfaces} are detected and registered to a map through point-to-line and point-to-plane constraints within an optimization framework. In LOAM, an IMU can be optionally used to de-skew the input point cloud and provide a prior motion estimate. LOAM extensions \change{can be found} to be used specifically on ground vehicles\change{~\cite{Shan2018}} or with solid-state LiDARs\change{~\cite{Lin2020,Li2021v2}}. More recently, a lightweight LOAM version named FLOAM~\cite{Wang2021} has been proposed. This is probably the closest work to our solution. In this respect, \appname{} introduces a simpler but more efficient pose optimization scheme that is based on LOAM edges, resulting into robust estimations, as it is shown in Section~\ref{sec:results}. Besides that, FLOAM uses downsampled global feature maps that are stored in 3D KD-trees. Our novel hashing-based mapping approach allows for a faster interaction with the map, especially regarding updates. \change{Rozenberszki et al.~\cite{Rozenberszki2021} also describes a LiDAR-based approach. However, in contrast to our solution, this method assumes the existence of a predefined map and relies on a place recognition method.} Finally, within this class of solutions, \cite{Yang2018,Demir2019} propose a data fusion scheme based on an Extended Kalman Filter (EKF).


Tightly-coupled methods fuse sensor data jointly, either through optimization~\cite{Ye2019,Shan2020} or filtering~\cite{Qin2020lins,Xu2020}. In this regard, Ye et al.~\cite{Ye2019} introduces LIOM, a tightly-coupled odometry and mapping approach which jointly minimizes LiDAR and IMU observations in a sliding window. Despite its good performance, it is computationally expensive, making difficult its use in practical situations. In a more recent work~\cite{Qin2020lins}, the same authors opt for an iterated Error-State Kalman Filter (ESKF), resulting into a faster solution. A recent work~\cite{Shan2020} introduces LIO-SAM as a new tightly-coupled method. In LIO-SAM, LiDAR-inertial odometry is stated as a factor graph, \change{allowing an easy incorporation of any type of observation as a constraint, such as loop closures, GPS samples or IMU measurements.} \change{In a very recent work~\cite{LeGentil2021}, the authors employ pre-integrated IMU readings to de-skew the input point cloud.} Unlike the approaches surveyed so far, we tackle the problem of pose estimation using solely a LiDAR.

\subsection{\change{Vision-based Solutions}}

\change{In the last decades, cameras have been extensively used for mapping and localization, specially motivated by the low cost of cameras, the constant increase in computing power and the richness of the sensor data provided. These solutions are typically classified into feature-based (indirect) and direct methods, according to the strategy employed for data association across images. On the one hand, feature-based solutions utilize a set of feature points to reconstruct the world around the sensor. In this regard, Klein and Murray introduced PTAM~\cite{klein2007parallel}, where tracking and mapping procedures run in parallel in two different threads. Following these ideas, Strasdat proposed a number of solutions~\cite{strasdat2010scale,strasdat2011double,strasdat2012visual} that served as a basis for ORB-SLAM~\cite{mur2015ORB,Mur2017ORBSLAM2,ORBSLAM3_TRO}, which can be considered as one of the most popular solutions in Visual SLAM. In \appname{}, we follow the computational approach adopted for these solutions, i.e. two threads run in parallel for feature extraction and pose estimation; additionally, a set of edge points are extracted from the point cloud.} 

\change{On the other hand, direct methods enable 3D reconstructions of the world of higher accuracy, since they use a larger amount of image pixels and minimize photometric errors for pose estimation. In this category, several solutions have been proposed, e.g. DTAM~\cite{Newcombe2011DenseTracking}, Kinect Fusion~\cite{Newcombe2011kinectFusion}, DVO~\cite{steinbrucker2011real}, Elastic Fusion~\cite{whelan2015ElasticFusionDS} and Kintinuous~\cite{kahler2016real}. However, they typically rely on a GPU to solve the pose estimation problem, which is out of the scope of this work. In comparison with vision-based approaches, LiDAR-based solutions are less affected by illumination changes, they can provide a full overview of the surroundings and can supply reliable range estimations.}


\change{A final class of solutions combine LiDAR and cameras~\cite{Zhang2015,Graeter2018,Seo2019}. The main objective of these methods is to obtain the advantages of both types of sensors, although at the expense of additional calibration and synchronisation processes that can significantly complicate their use together.}

\subsection{\change{Map Indexing}}


As mentioned above, establishing a set of correspondences between the input scan and a map is of prime importance for efficient pose estimation. Some authors have opted for indexing the points of a global map using a tree-based approach~\cite{Zhang14}, although usually these solutions do not scale well. In our work, the global map is devised as a disjoint partition of the 3D space, and, inspired by other approaches~\cite{Lin2019,Zhao2021}, it is indexed using a hashing scheme. An alternative for fast data association is building a local map from a sliding window~\cite{Ye2019,Shan2020}, instead of matching directly to a global map, but this option tends to discard useful correspondences. In this matter, \appname{} also introduces an adaptive local map mechanism, which can be seen as an alternative to the classical local mapping paradigm.

\section{System Overview}
\label{sec:sysoverview}

\begin{figure}[tb]
\centering
\includegraphics[width=1.0\columnwidth]{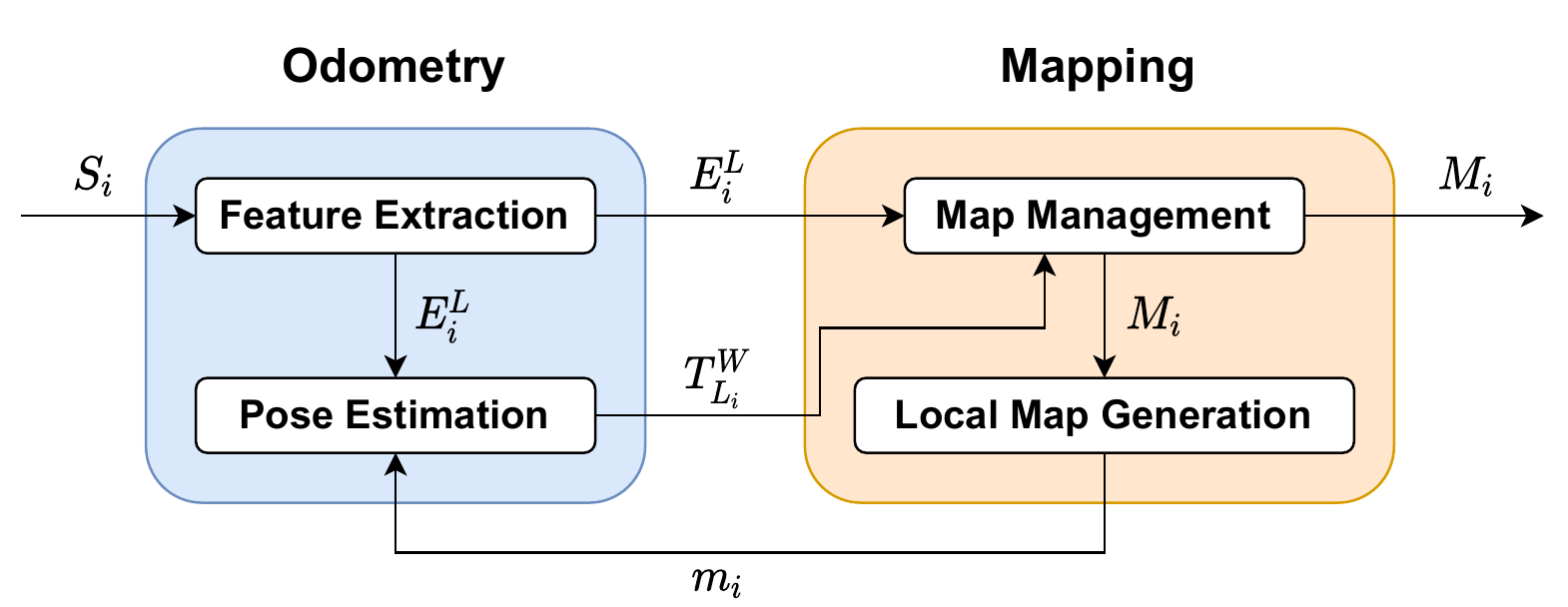}
\caption{Overview of \appname{}.}
\label{fig:liodomarch}
\end{figure}

For a start, we define a \textit{sweep} as a set of 360-degree 2D scans. A sweep received at time $i$ is denoted as $S_i$. We also define two coordinate systems: (1) $L$, the \textit{LiDAR} coordinate system, which is a frame attached to the geometric center of the sensor; and (2) $W$, the world coordinate system, which coincides with $L$ at the beginning. We denote by $T^A_B \in SE(3)$ the transformation that maps a point $p^B \in \mathbb{R}^3$ expressed in $B$ to a point $p^A \in \mathbb{R}^3$ expressed in $A$. The rotation matrix and the translation vector of $T^A_B$ are respectively denoted by $R^A_B \in SO(3)$ and $t^A_B \in \mathbb{R}^3$.


Figure~\ref{fig:liodomarch} illustrates \appname{}. As in other works~\cite{Zhang14,Ye2019}, our solution consists of two main components, odometry and mapping, which run concurrently: while the odometry module (Sec.~\ref{sec:odom}) computes a set of features \change{$E^L_i$ (LOAM edges)} from $S_i$ and estimates the current pose of the LiDAR $T^W_{L_i}$, the mapping module (Sec.~\ref{sec:mapping}) registers the resulting edges to a global map \change{$M_i$} and generates an adaptive local map \change{$m_i$} to be employed in the subsequent pose estimation step.

\section{LiDAR Odometry}
\label{sec:odom}


The LiDAR odometry module \change{is organized into two synchronized execution threads that, hence, decouples} feature extraction from pose estimation. Both are described next.

\subsection{Feature Extraction}
\label{sec:fext}

For a start, each sweep $S_i$ is divided into its different scans, discarding at the same time those points whose range do not fall within a certain interval $[r_\text{min}, r_\text{max}]$, which has to be configured accordingly to the sensor operating and noise characteristics. Each pre-processed scan is next considered, selecting a number of key points to reduce the computational requirements. In this work, we make use of LOAM edge features, given the utility they have shown in other works and their simpler computation~\cite{Zhang14,Shan2018,Wang2021,Wang2020}. Besides, to select the best features, in \appname{} we also calculate a local curvature measure $c_j$ for each point $p^{L_i}_j$~\cite{Zhang14} as follows:
\begin{equation}
    c_j = \frac{1}{|\change{\Omega}| \cdot || p^{L_i}_j  ||} \sum_{k \in \change{\Omega}, k \neq j} || p^{L_i}_j - p^{L_i}_k||\,,
    \label{eq:curv}
\end{equation}
being \change{$\Omega$ a} set of consecutive (\change{i.e. co-planar}) points \change{in the vicinity} of $p^{L_i}_j$ \change{belonging to} the same scan of \change{the sweep} $S_i$. \change{In this way, we take advantage of the typical larger intra-beam resolution against the inter-beams resolution, to get more distinctive features~\cite{Wang2021}.} Moreover, to distribute edges throughout the environment, a scan is further divided into equally-sized sectors, and a maximum number of edges is set for every sector. Unlike~\cite{Zhang14}, we split each scan into 8 sectors and choose a maximum of 10 edges per sector after sorting them in decreasing order of curvature $c$. Furthermore, the selection applies non-maxima suppression, i.e. a point is chosen as an edge if none of its neighbours has been already selected. The result of this procedure is a set of edges $E^L_i$ chosen from sweep $S_i$.
\subsection{Pose Optimization}
\label{sec:popt}

\begin{algorithm}[tb]
\caption{LiDAR Odometry}
\label{alg:lidarodom}
\begin{algorithmic}[1]

\Require{$S_i$, $T^W_{L_{i-1}}$, $T^W_{L_{i-2}}$, $m_{i-1}$}
\Ensure{$E^L_i$, $T^W_{L_{i}}$}

\State $E^L_i$ $\leftarrow$ set of edges from $S_i$
\State $T^{\,W}_{L_{i}}$ $\leftarrow$ $\widehat{T}^{\,W}_{L_{i}} \equiv$ initial transformation estimate \hfill [Eq. (\ref{eq:inittrans})]
\For{$n$ iterations}
    \For{$p^{L_i}_j \in E^L_i$}
        \State $p^{W}_j \leftarrow T^W_{L_{i}} p^{L_i}_j$ \hfill [Eq. (\ref{eq:p2world})]
        \State $N(p^{W}_j) \leftarrow$ 5 NN of $p^{W}_j$ in $m_{i-1}$ 
        \If{$N(p^{W}_j)$ is a line}
            \State Compute $d_e(p^{W}_j, l(p^{W}_j))$ \hfill [Eq. (\ref{eq:p2ldist})]
            \State Compute residual $\varrho_e(p^{W}_j, l(p^{W}_j))$ \hfill [Eq. (\ref{eq:residual})]
            \State Add residual $\varrho_e(p^{W}_j, l(p^{W}_j))$ 
            \State \quad to the optimization problem
        \EndIf
    \EndFor
    \State Optimize pose $T^W_{L_i}$ \hfill [Eq. (\ref{eq:optproblem})]
\EndFor
\end{algorithmic}
\end{algorithm}

Let us consider the transformation $T^W_{L_{i}}$ from the LiDAR at time $i$ to the world. Then, every point $p^{L_i}_j \in E^L_i$ projects into the world frame $W$ as:
\begin{equation}
    p^{W}_j = T^W_{L_{i}} p^{L_i}_j = R^W_{L_i} p^{L_i}_j + t^{\,W}_{L_i}\,,
    \label{eq:p2world}
\end{equation}
being $R^W_{L_i}$ and $t^{W}_{L_i}$ the respective rotation matrix and translation vector of $T^{W}_{L_i}$. We denote the set of transformed edges as $E_{i}^{W}$. Subsequently, a set of point-to-line correspondences between $E_i^W$ and a local map are computed for pose estimation. We have opted for this solution rather than using, for instance, a global map, because it turns out to be more computationally stable as more frames are processed. In \appname{}, that local map is not built after pose estimation in a sequential way as in~\cite{Ye2019,Shan2020}, but it is built concurrently with pose estimation by the mapping module (see Section~\ref{sec:mapping}). 

Let us assume now the existence of a local map $m_{i-1}$ at time $i-1$, which is a subset of the global map $M_{i-1}$. This map $m_{i-1}$ contains the points in $M_{i-1}$ closest to the LiDAR according to the latest pose estimate $T^W_{L_{i-1}}$. For each point $p^{W}_j \in E_i^W$, we obtain the $k$ nearest points in $m_{i-1}$, where $k = 5$ in this work. \change{We now} denote this set as $N(p^{W}_j$) and the $n$-th nearest neighbour of $p^{W}_j$ as $N_{n}(p^{W}_j$). \change{Next, we} assess whether points in $N(p^{W}_j$) are aligned by analyzing their scatter matrix~\cite{Lin2020}. If the largest eigenvalue of this matrix is, at least, three times the second largest eigenvalue, we consider that a valid point-to-line correspondence can be established between $p^{W}_j$ and the line $l(p^{W}_j)$ resulting from $N_{1}(p^{W}_j$) and $N_{2}(p^{W}_j$). We then calculate the point-to-line distance $d_e$ as
\begin{equation}
    d_e(p^{W}_j, l(p^{W}_j)) = \frac{\left\|\left(p^{W}_j - N_{1}(p^{W}_j)\right) \times N_{12} \right\|}{\|\,N_{12}\,\|}\,,
    \label{eq:p2ldist}
\end{equation}
with $N_{12} = N_{1}(p^{W}_j) - N_{2}(p^{W}_j)$. \change{At this point, it is worth noting that increasing $k$ leads to higher computational times for calculating the scatter matrix, without necessarily ensuring additional insight on whether the nearby points lie on a line in a local sense.}

\appname{}, as an odometer, optimizes only the current pose of the LiDAR $T^W_{L_i}$. Within the optimization framework, each correspondence provides a constraint between $T^W_{L_i}$ and the local map $m_{i-1}$, whose residual $\varrho_e$ is computed as:
\begin{equation}
    \varrho_e(p^{W}_j, l(p^{W}_j)) = \omega_j \, d_e(p^{W}_j, l(p^{W}_j))\,,
    \label{eq:residual}
\end{equation}
where $\omega_j$ is a weighting term \change{defined} as:
\begin{equation}
    \omega_j = 1 - \frac{r_j - r_\text{min}}{r_\text{max} - r_\text{min}}\,,
    \label{eq:weight}
\end{equation}
being $r_j$ the range returned by the LiDAR for edge $p^{L_i}_j$. The rationale behind this factor is that LiDARs tend to decrease their accuracy at longer distances and, thus, we give more importance to correspondences established at closer distances. We then compute the optimal transformation $T^W_{L_i}$ as the minimizer of the loss function $J(\widetilde{T}^{\,W}_{L_i}, \Upsilon)$:
\begin{align}
    J(\widetilde{T}^{\,W}_{L_i}, \Upsilon) &= \frac{1}{2} \sum_{j \in \Upsilon} \rho \Big(\left\| \varrho_e\left(\widetilde{T}^W_{L_{i}} p^{L_i}_j, l\left(\widetilde{T}^W_{L_{i}} p^{L_i}_j\right)\right) \right\|^2\Big)
    \nonumber
    \\
    T^W_{L_i} &= \underset{\widetilde{T}^{\,W}_{L_i}}{\min}\ J(\widetilde{T}^{\,W}_{L_i}, \Upsilon)
    \label{eq:optproblem}
\end{align}
where $\Upsilon$ is the set of correspondences established between $E_i^W$ and the local map $m_{i-1}$, and $\rho$ is a Huber loss function to reduce the influence of outliers. The system of non-linear equations is solved by means of the Levenberg-Marquardt algorithm using the Ceres Solver library\cite{ceres}, using \change{the} transformation $\widehat{T}^{\,W}_{L_i}$ as initial guess:
\begin{align}
    \widehat{T}^W_{L_i} &= T^W_{L_{i-1}} \widehat{T}^{L_{i-1}}_{L_i} \nonumber \\
                        &= T^W_{L_{i-1}} T^{L_{i-2}}_{L_{i-1}}
                         = T^W_{L_{i-1}} \left(T_{L_{i-2}}^{W}\right)^{-1} T^W_{L_{i-1}}\,,
    \label{eq:inittrans}
\end{align}
i.e. we assume the same motion as for the previously estimated pose. Although \appname{} deals only with LiDAR data, it is clear that any additional motion estimate, e.g. from an IMU, can be incorporated at this point. 

The full LiDAR odometry procedure is stated algorithmically in Alg.~\ref{alg:lidarodom}. In our experiments, 1 or 2 refining iterations are enough, i.e. $n=2 \text{ or } 3$ at line 3 of Alg.~\ref{alg:lidarodom}.

\section{LiDAR Mapping}
\label{sec:mapping}

The registration of the extracted edges $E_i^{L}$ on the global map $M_i$ is performed by the mapping module using the last optimized pose $T^W_{L_i}$. This module also generates the corresponding local map $m_i$ as described next.

\subsection{Map Representation}
\label{sec:map}

Given the high frequency at which the map must be accessed, the type of data structure chosen to represent 3D space becomes crucial for fast operation. A single KD-tree has been typically used to this end~\cite{Zhang14}. However, this option presents several drawbacks: on the one hand, the full tree tends to change as points are added or deleted to/from the tree, and, on the other hand, the KD-tree performance decreases as more points need to be managed~\cite{Zhang14}. To overcome these issues, in \appname{} we introduce an efficient hashing data structure for representing the map, taking inspiration from other recent works~\cite{Lin2019,Zhao2021}. To be more specific, the 3D space is partitioned into a set of disjoint cuboids of a fixed size that we name \textit{cells}. A cell $C_j$ is represented by its geometric center, denoted by ($c_{jx}, c_{jy}, c_{jz}$), and includes all 3D points whose coordinates fall into its limits. We define a map at time $i$ as $M_i = \{\mathbb{H}_i, \mathbb{C}_i\}$, where $\mathbb{H}_i$ is a hash table and $\mathbb{C}_i$ is the set of existing cells up to time $i$. The table $\mathbb{H}_i$ allows us to rapidly \change{get access} to a specific cell $C_j$ using a hash function of its coordinates, defined by:
\begin{equation}
    H(C_j) = (c_{jx} \oplus (c_{jy}\,{\scriptstyle <<}\,1)) \oplus (c_{jz}\,{\scriptstyle <<}\,2)\,,
    \label{eq:hashfun}
\end{equation}
where $\oplus$ and $\,{\scriptstyle <<}$ are, respectively, the bitwise XOR and the left shift operators. This function has been selected in order to minimize, as much as possible, hash collisions. That is to say, if bits of a binary word have roughly 50\% chance of being 0 or 1, i.e. as randomly distributed as possible, the bitwise XOR between such binary words results into another word also following a random distribution. Furthermore, since the bitwise XOR is a symmetric operation, the order of the elements in the hash code is lost. To break this symmetry, we use the shift operator, at a limited computational cost.

\subsection{Map Updates}
\label{sec:mapupdate}

\begin{algorithm}[tb]
\caption{LiDAR Mapping}
\label{alg:mapping}
\begin{algorithmic}[1]

\Require{$E^L_i$, $T^W_{L_{i}}$, \change{$M_{i-1}$}}
\Ensure{$M_{i}$, $m_i$}

\State \change{$M_i$ $\leftarrow$ $M_{i-1}$}
\For{$p^{L_i}_j \in E^L_i$}
    \State $p^{W}_j \leftarrow T^W_{L_{i}} p^{L_i}_j$ \hfill [Eq. (\ref{eq:p2world})]
    \State $C_q$ $\leftarrow$ cell where $p^{W}_j$ should be \hfill [Eq. (\ref{eq:cellpoint})]
    
    \If{$H(C_q) \not\in \mathbb{H}$}
        \State Create new cell $C_n$ using $C_q$ coordinates
        \State Add $p^{W}_j$ to $C_n$
        \State Update $M_i$ adding $C_n$ to $\mathbb{C}$
        \State Update $M_i$ adding $H(C_q)$ to $\mathbb{H}$
    \Else
        \State Retrieve cell $C_q$ using $H(C_q)$
        \State Update $M_i$ adding $p^{W}_j$ to $C_q$
        \If{$C_q$ has more points than $\tau$}
            \State Update $M_i$ filtering $C_q$ using a 3D voxel grid
        \EndIf
    \EndIf
\EndFor

\State $C_{L_i}$ $\leftarrow$ cell where the LiDAR should be \hfill [Eq. (\ref{eq:cellpoint})]
\State $m_i$ $\leftarrow$ $\emptyset$
\For{$C_i \in$ Neighbours of $C_{L_i}$ in $M_i$}
    \State $m_i$ $\leftarrow$ $m_i \cup C_i$
\EndFor
\change{
\For{$j \in \{1, 2, 3\}$}
    \State $m_i$ $\leftarrow$ $m_i \cup E^{W}_{i-j}$
\EndFor
}
\end{algorithmic}
\end{algorithm}

In \appname{}, map updates are performed once per sweep, being the set of edges $E_i^{L}$, extracted from $S_i$, and the last optimized transformation $T^W_{L_i}$ the input data. Unlike other approaches~\cite{Zhang14}, where the raw point cloud is used for mapping, in our approach, the map is built using directly the edges to speed up the mapping procedure, resulting into more sparse maps. Initially, every point $p^{L_i}_j \in E_i^{L}$ is transformed to world coordinates using $T^W_{L_i}$ and (\ref{eq:p2world}). Next, for each point $p^{W}_j = (x, y, z)$, we compute the geometric center of the cell $C_q$ in which the point should be stored as:
\begin{equation}
    \begin{bmatrix}
    c_{qx} \\ \\
    c_{qy} \\ \\
    c_{qz}
    \end{bmatrix}
    = 
    \begin{bmatrix}
    \left\lfloor x / s_{xy} \right\rfloor \, s_{xy} + \frac{1}{2}\,s_{xy}\\ \\
    \left\lfloor y / s_{xy} \right\rfloor \, s_{xy} + \frac{1}{2}\,s_{xy}\\ \\
    \left\lfloor z / s_{z}  \right\rfloor \, s_{z}  + \frac{1}{2}\,s_{z}\\
    \end{bmatrix}\,,
    \label{eq:cellpoint}
\end{equation}
where $s_{xy}$ and $s_z$ are the metric cell sizes for the corresponding dimension. We next check if the cell $C_q$ is already in the map by querying the hash table $\mathbb{H}$ using the key $H(C_q)$. If this is the case, the point is added to the existing cell. Otherwise, a new cell $C_n$ is created with point $p^{W}_j$ as seed, to be added next to $\mathbb{C}$ and indexed on $\mathbb{H}$ by $H(C_n)$. Finally, modified cells exceeding a certain number of points are filtered using a 3D voxel grid. Note that our data structure allows us to rapidly update just the required areas of the environment, avoiding the update of the whole map at each iteration. This fact contributes to speed up the mapping procedure, as will be shown in the experiments.

\subsection{Adaptive Local Map Computation}
\label{sec:localmap}

Lastly, the mapping module generates a local map $m_i$, which contains the points of $M_i$ within a certain range from the current LiDAR pose. Assuming a moderate motion between two consecutive sweeps, these points are enough to find correspondences for the next pose estimation step. To build the local map, we first retrieve the cell $C_{L_i}$ where the LiDAR is located at that moment using its current position $T^W_{L_i}$ and (\ref{eq:cellpoint}). Next, assuming a 3D grid arranged over $M_i$, neighbouring cells of $C_{L_i}$ up to a certain distance are further retrieved from $M_i$, and their corresponding points are merged to form the local map $m_i$. This operation results to be very fast due to the proposed hashing structure. Points on $m_i$ are finally organised into a KD-tree to speed up nearest neighbour search. Note that this tree is very simple, as it just contains a small subset of the total map points, in contrast to managing the whole global map~\cite{Zhang14}.

On the other side, we refer to this local map as \textit{adaptive} since it always covers a specific area of the environment, contrary to a local map built by aggregation of a sliding window~\cite{Ye2019,Shan2020}. Besides, it provides us with correspondences with revisited areas of the environment in a natural way. Additionally, the availability of $m_i$ avoids us to search for correspondences against the whole map, as done by other solutions~\cite{Zhang14}. Finally, to avoid reduced amounts of points from unexplored areas, we always add the last three sweeps to $m_i$. The complete mapping procedure is outlined in Alg.~\ref{alg:mapping}.

\section{Experimental Results}
\label{sec:results}

In this section, we report on the results of several experiments conducted to evaluate \appname{}, including a comparison with other solutions. A laptop featuring an Intel Core i7-10750H @2.6Ghz, 16 GB RAM has been used in all cases.

\subsection{Methodology}
\label{sec:method}

We validate our approach using the KITTI odometry benchmark~\cite{Geiger2012}, as usual in the field~\cite{Wang2021,Li2021,Zheng2021}. This dataset consists of 22 sequences collected using a Velodyne HDL-64E sensor. Eleven of these sequences include GPS poses that can be used as ground truth. The average translational ($\%$) and rotational ($\text{deg}/100\text{m}$) errors are adopted in the following as main performance metrics. We additionally consider the Absolute Trajectory Error (ATE), although it rather focuses on the global consistency of the whole trajectory and thus it is more appropriate for SLAM systems.

To further validate \appname{}, we compare it with other pure LiDAR-based odometry and also with SLAM solutions, namely FLOAM~\cite{Wang2021}, ISC-LOAM~\cite{Wang2020} and LeGO-LOAM~\cite{Shan2018}. We are aware of the existence of recent fusion-based~\cite{Ye2019,Shan2020} or even semantic-aided~\cite{Li2021} solutions. They are not considered in this evaluation since, in contrast to our method, they imply additional complexities, such as synchronization and calibration procedures or increasing computational resources.

\subsection{Algorithm Configuration}

\begin{table}[tb]
\caption{\change{Parameters values and section where they are defined.}}
\begin{center}
\begin{tabular}{r|l||r|l}
\hline
$r_\text{min}$ (Sec.~\ref{sec:fext}) & 3.0 & $k$ (Sec.~\ref{sec:popt}) & 5\\
$r_\text{max}$ (Sec.~\ref{sec:fext}) & 75.0 & $S_{xy}$ (Sec.~\ref{sec:mapupdate}) & 25\\
Scan sectors (Sec.~\ref{sec:fext}) & 8 & $S_{z}$ (Sec.~\ref{sec:mapupdate}) & 20\\
Edges per sector (Sec.~\ref{sec:fext}) & 10 & Sweeps in $m_i$ (Sec.~\ref{sec:localmap}) & 3 \\
\hline
\end{tabular}
\end{center}
\label{tab:params}
\end{table}


\change{
\appname{} has been run several times against the K05 sequence, fine-tuning them towards increasing the localization accuracy, in order to find a suitable set of parameters. The values obtained for the most relevant parameters can be found in Table~\ref{tab:params}. They have been kept constant in all the remaining experiments, what, given the performance achieved, shows that this configuration works reasonably well and is able to tolerate different operating conditions: two different LiDAR devices, different processors and different datasets taken from different environments. This set-up represents, thus, a good trade-off between performance and accuracy. Nonetheless, those parameters with an impact on the computational complexity could need specific modifications depending on the available computational resources in order to keep response times at a reasonable value.}

\subsection{Odometry Performance}
\label{sec:perfeval}

\begin{table*}[tb]
\scriptsize
\caption{Average translational and rotational errors for the KITTI odometry benchmark. \\
Best results are shown in bold red and second best in blue.}
\begin{center}
\begin{tabularx}{\textwidth} { 
  >{\centering\arraybackslash}X |
  >{\centering\arraybackslash}X >{\centering\arraybackslash}X
  >{\centering\arraybackslash}X >{\centering\arraybackslash}X ||
  >{\centering\arraybackslash}X >{\centering\arraybackslash}X
  >{\centering\arraybackslash}X >{\centering\arraybackslash}X}
& \multicolumn{4}{c||}{\textbf{Translational Error ($\%$)}} & \multicolumn{4}{c}{\textbf{Rotational Error (deg/100m)}}\\
\midrule
& \textbf{FLOAM} & \textbf{ISC-LOAM} & \textbf{LeGO} & \textbf{Ours} & \textbf{FLOAM} & \textbf{ISC-LOAM} & \textbf{LeGO} & \textbf{Ours}\\
\midrule
\textbf{K00} & \textcolor{blue}{0.861} & 1.020	& 2.170	 & \textcolor{red}{\textbf{0.857}} & \textcolor{blue}{0.349} & 0.420 & 1.050 & \textcolor{red}{\textbf{0.348}}\\
\textbf{K01} & \textcolor{blue}{1.309} & 2.920	& 13.400 & \textcolor{red}{\textbf{1.301}} & \textcolor{red}{\textbf{0.128}} & 0.630 & 1.020 &  \textcolor{blue}{0.129} \\
\textbf{K02} & \textcolor{blue}{0.952}	& 1.670	& 2.170	 & \textcolor{red}{\textbf{0.947}} & \textcolor{blue}{0.310} & 0.540 & 1.010 & \textcolor{red}{\textbf{0.309}}\\
\textbf{K03} & 1.267	& \textcolor{red}{\textbf{1.150}}	& 2.340	 & \textcolor{blue}{1.262} & \textcolor{blue}{0.227} & 0.720 & 1.180 & \textcolor{red}{\textbf{0.226}}\\
\textbf{K04} & 1.417	& 1.500	& \textcolor{red}{\textbf{1.270}} & \textcolor{blue}{1.411} & \textcolor{blue}{0.010} & 0.560 & 1.010 & \textcolor{red}{\textbf{0.009}}\\
\textbf{K05} & 0.835	& \textcolor{red}{\textbf{0.810}}	& 1.280	 & \textcolor{blue}{0.834} & \textcolor{blue}{0.360} & 0.370 & 0.740 & \textcolor{red}{\textbf{0.359}}\\
\textbf{K06} & 0.835	& \textcolor{red}{\textbf{0.760}} & 1.060	 & \textcolor{blue}{0.834} & \textcolor{blue}{0.332} & 0.410 & 0.630 & \textcolor{red}{\textbf{0.331}}\\
\textbf{K07} & 0.883	& \textcolor{red}{\textbf{0.560}} & 1.120 & \textcolor{blue}{0.881} & 0.617 & \textcolor{red}{\textbf{0.430}} & 0.810 & \textcolor{blue}{0.614}\\
\textbf{K08} & \textcolor{blue}{0.869}	& 1.200	& 1.990	 & \textcolor{red}{\textbf{0.864}} & \textcolor{blue}{0.332} & 0.500 & 0.940 & \textcolor{red}{\textbf{0.331}}\\
\textbf{K09} & \textcolor{blue}{1.033}	& 1.400	& 1.970	 & \textcolor{red}{\textbf{1.029}} & \textcolor{red}{\textbf{0.317}} & 0.590 & 0.980 & \textcolor{blue}{0.318}\\
\textbf{K10} & \textcolor{blue}{1.203}	& 1.870	& 2.210	 & \textcolor{red}{\textbf{1.196}} & \textcolor{red}{\textbf{0.287}} & 0.620 & 0.920 & \textcolor{blue}{0.288}\\
\midrule
\textbf{Average} & \textcolor{blue}{1.042} & 1.351 & 2.816 & \textcolor{red}{\textbf{1.038}} & \textcolor{blue}{0.297} & 0.526 & 0.935 & \textcolor{red}{\textbf{0.296}}\\
\end{tabularx}
\end{center}
\label{tab:errors}
\end{table*}

Table~\ref{tab:errors} summarizes the results obtained in terms of translational and rotational errors. Results for FLOAM were obtained by ourselves using its open source implementation, while results for ISC-LOAM and LeGo-LOAM are directly reported from, respectively, \cite{Li2021} and~\cite{Zheng2021}. As can be observed, \appname{} achieves competitive results in all sequences in terms of translation error. This can be observed even in sequences comprising loop closures, such as K05, K06 and K07, where our approach achieves the second best results, sometimes very close to complete SLAM solutions like ISC-LOAM. We obtain, on average, 1.038\% drift in translation, outperforming the other solutions in this matter. Regarding rotation error, again our solution leads to the lowest errors in most of the sequences. On average, the rotational error of \appname{} is 0.296\% deg / 100m, which represents again the lowest average error.

\begin{table}[tb]
\small
\caption{Absolute Trajectory Error (m) for the KITTI dataset. Best results are shown in bold red and second best in blue.}
\begin{center}
\begin{tabularx}{\columnwidth} { 
  >{\centering\arraybackslash}X |
  >{\centering\arraybackslash}X >{\centering\arraybackslash}X >{\centering\arraybackslash}X >{\centering\arraybackslash}X}
& \textbf{FLOAM} & \textbf{ISC-LOAM} & \textbf{LeGO} & \textbf{Ours}\\
\midrule
\textbf{K00} & \textcolor{blue}{5.137} & \textcolor{red}{\textbf{1.600}} & 6.300 & 7.135\\
\textbf{K02} & \textcolor{blue}{9.294} & \textcolor{red}{\textbf{4.770}} & 14.700 & 9.754\\
\textbf{K05} & 2.546 & \textcolor{blue}{2.490} & 2.800 & \textcolor{red}{\textbf{0.322}}\\
\textbf{K06} & \textcolor{blue}{0.934} & 1.030 & \textcolor{red}{\textbf{0.800}} & 0.956\\
\textbf{K07} & \textcolor{red}{\textbf{0.498}} & \textcolor{blue}{0.560} & 0.700 & 1.518\\
\textbf{K08} & \textcolor{blue}{4.344} & 4.880 & \textcolor{red}{\textbf{3.500}} & 4.592\\
\textbf{K09} & 2.144 & 2.310 & \textcolor{blue}{2.100} & \textcolor{red}{\textbf{0.470}}\\
\midrule
\textbf{Average} & 3.557 & \textcolor{red}{\textbf{2.520}} & 4.414 & \textcolor{blue}{3.535}\\
\end{tabularx}
\end{center}
\label{tab:errorsATE}
\end{table}

\begin{figure*}[tb]
    \begin{center}
        \begin{tabular}{@{\hspace{0.0mm}}c@{\hspace{0.0mm}}c@{\hspace{0.0mm}}}
        \includegraphics[width=0.45\textwidth]{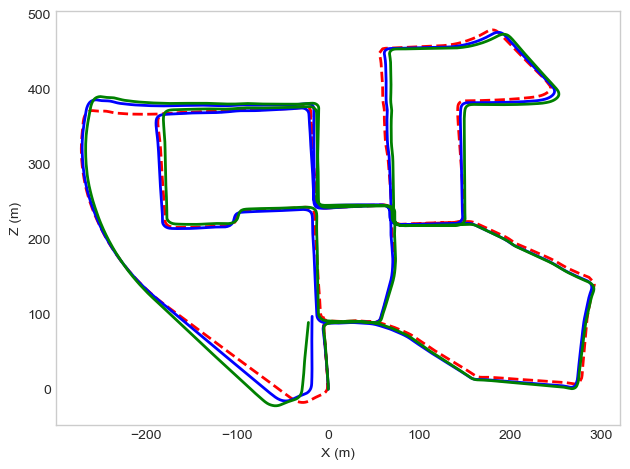} &
        \includegraphics[width=0.45\textwidth]{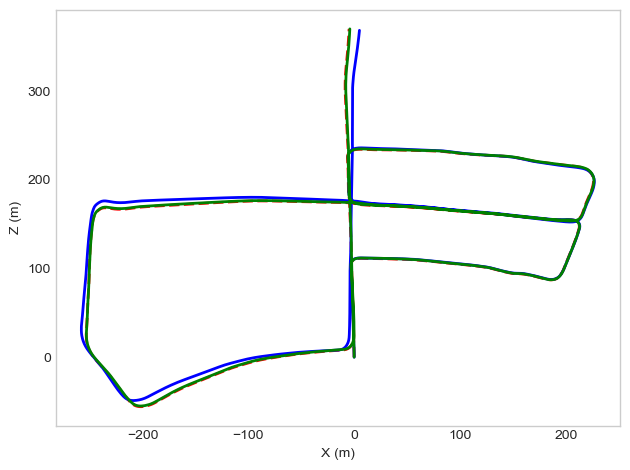} \\
        K00 & K05 \\
        \includegraphics[width=0.45\textwidth]{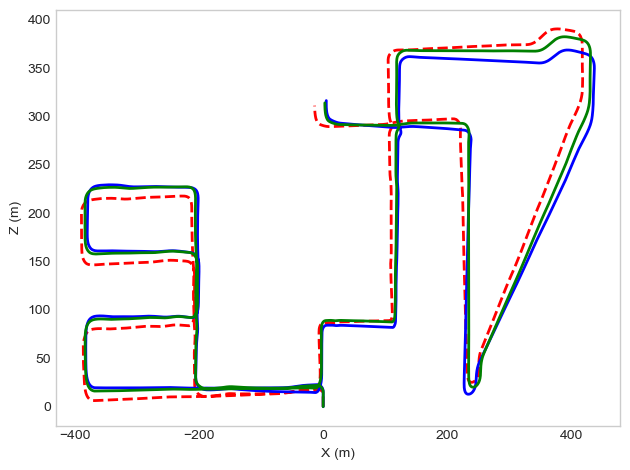} &
        \includegraphics[width=0.45\textwidth]{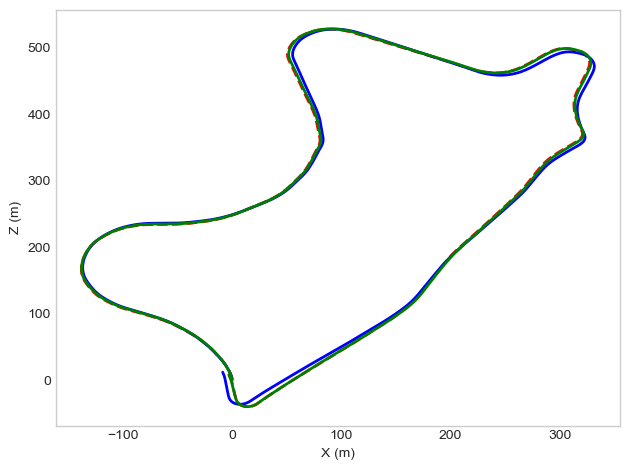} \\
        K08 & K09
        \end{tabular}
    \caption{Examples of trajectories estimated for some sequences of the KITTI odometry benchmark. The ground truth is shown as a red dashed line, while FLOAM and \appname{} estimates are respectively shown as blue and green lines.}
    \label{fig:paths}
    \end{center}
\end{figure*}


Table~\ref{tab:errorsATE} reports on the ATE for the KITTI sequences that contain loop closures. Again, results for FLOAM were obtained by ourselves, while results for ISC-LOAM and LeGo-LOAM are reported from, respectively,~\cite{Li2021} and~\cite{Yokozuka2021}. \appname{} again achieves competitive results in all sequences despite it is actually a pure odometry system and, therefore, does not take any advantage from global map optimization nor from loop closures. \change{In contrast to other solutions based on a sliding window policy, our pose estimation procedure makes use of an adaptive local map covering, in all directions, the surroundings of the LiDAR at its current position. This strategy allows our approach to find associations/matchings in a wider area than just the one covered by the previous frames, hence effectively reducing the pose estimation error.} On average, the ATE for \appname{} results to be 3.535 m, which represents the second best performance among the different methods considered. By way of illustration, Fig.~\ref{fig:paths} shows the resulting trajectory estimates from our approach and from FLOAM for several KITTI sequences.

\begin{figure}[tb]
\centering
\includegraphics[width=1.0\columnwidth,]{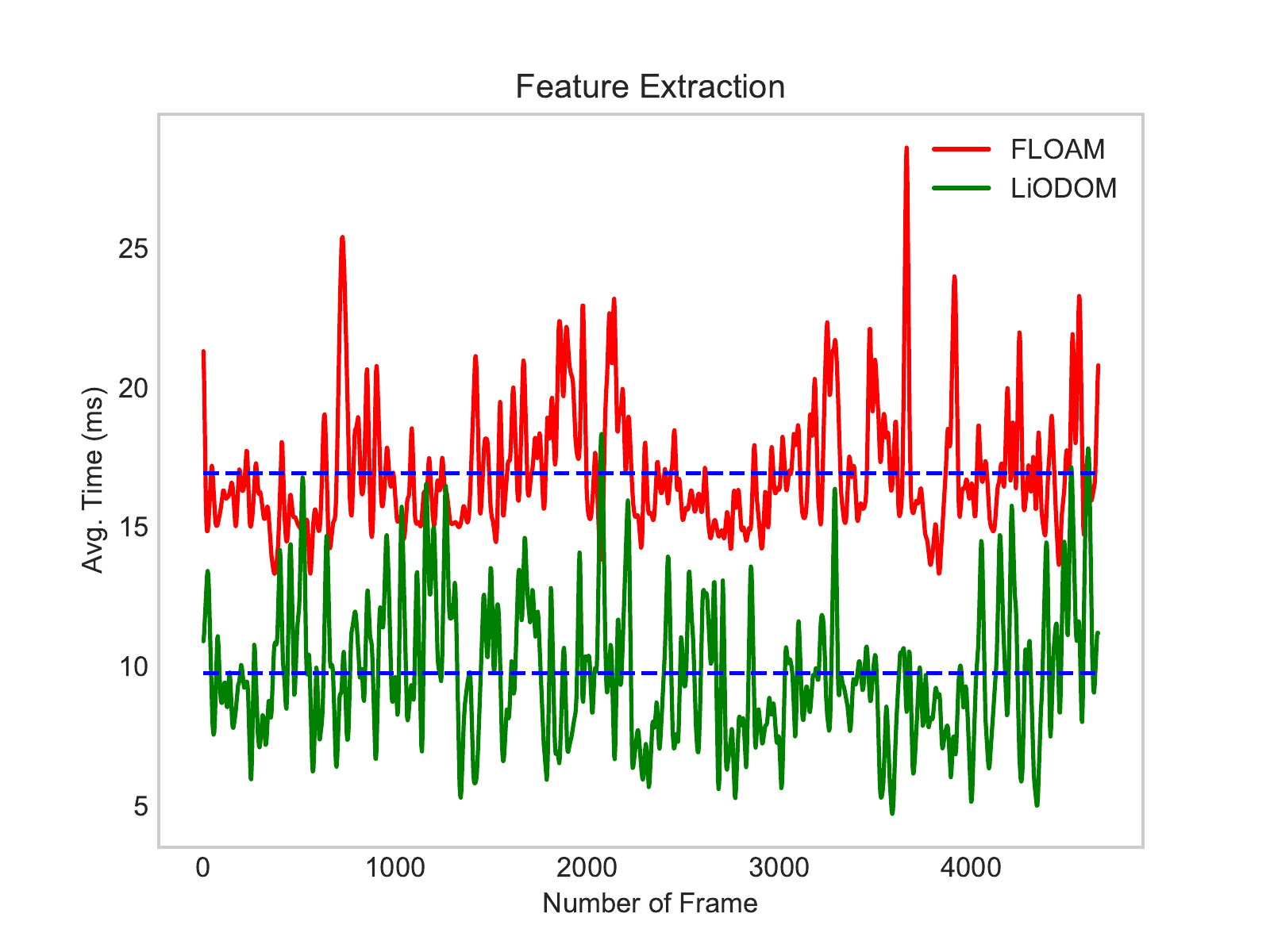}
\caption{Average processing times of the feature extraction stages of FLOAM and \appname{}. The blue dashed line corresponds to the mean value for each case.}
\label{fig:times_feat}
\end{figure}

\begin{figure}[tb]
\centering
\includegraphics[width=1.0\columnwidth,]{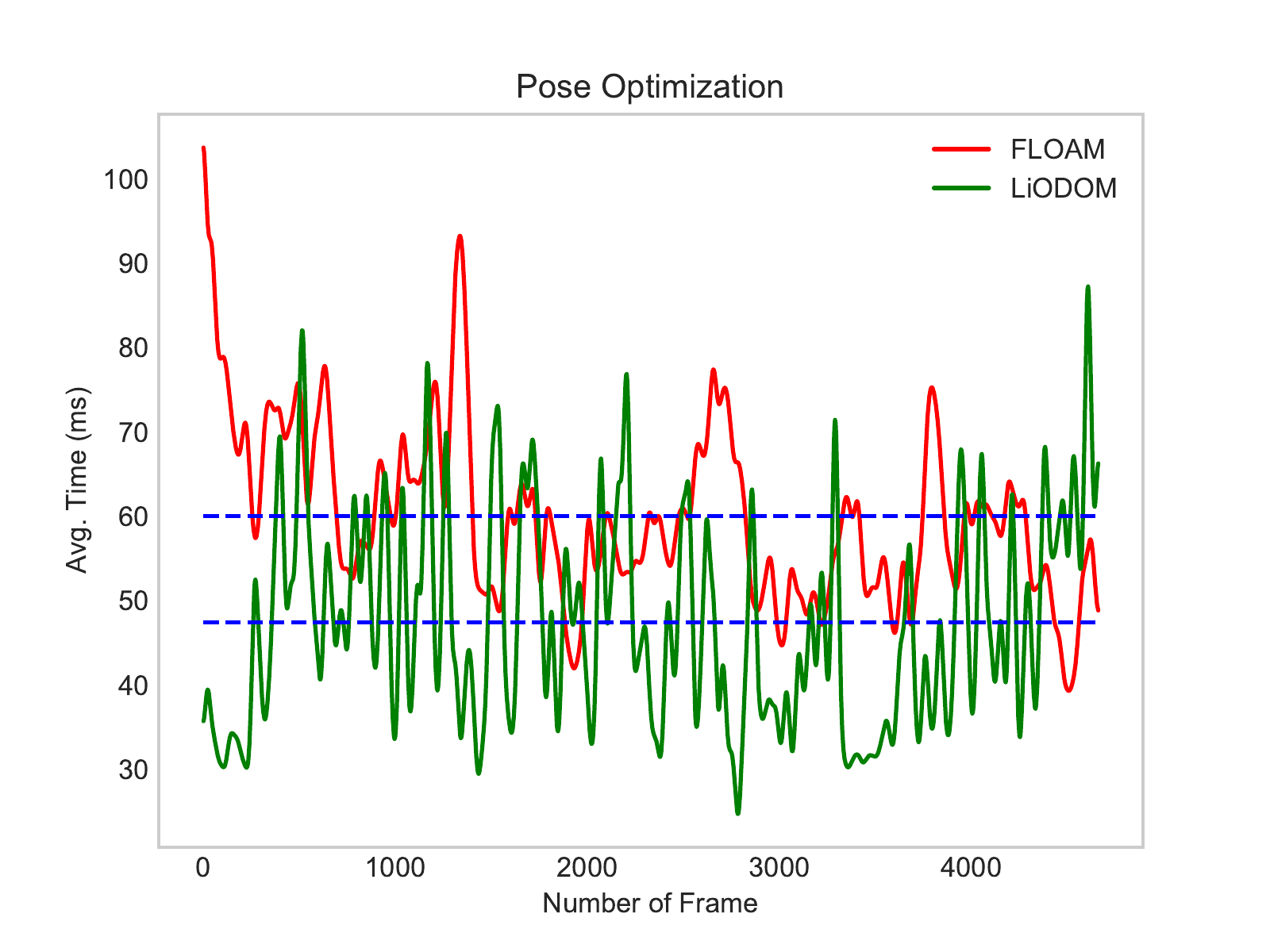}
\caption{Average processing times of the pose optimization stages of FLOAM and \appname{}. The blue dashed line corresponds to the mean value for each case.}
\label{fig:times_optim}
\end{figure}


In \change{the following}, we analyze the computational complexity of \appname{}. To this end, we choose dataset K02, the largest dataset considered in this work. Average processing times for every odometry stage can be found in Fig.~\ref{fig:times_feat} and Fig.~\ref{fig:times_optim}, as they are the stages that operate online within a real platform. The capabilities of the mapping module, which is executed as a standalone procedure, are evaluated in the next section. In both figures, we compare \appname{} with FLOAM as a representative of the solutions based on LOAM edges and surfaces as features~\cite{Zhang14,Shan2018,Wang2021,Wang2020} \change{(notice that the other solutions considered in this section can also be regarded as derivatives of LOAM)}. Regarding feature extraction, Fig.~\ref{fig:times_feat} shows that \appname{} presents lower extraction times than FLOAM. This stage takes, on average, 10.36 ms in contrast to the 16.92 ms required by FLOAM. Alternatively, Fig.~\ref{fig:times_optim} shows pose optimization times for each, including the search for correspondences within the respective maps. As can be observed, \appname{} converges, in general, faster than FLOAM, taking, on average, 48.73 ms in contrast to 60.12 ms. Considering the measured feature extraction and pose optimization times, the resulting overall processing times are 59.09 ms per frame for \appname{} and 77.04 ms per frame for FLOAM, which means, respectively, frame rates of around 17 Hz and 13 Hz. This means that \appname{} is about 1.3 times faster than FLOAM.

\begin{figure}[tb]
\centering
\includegraphics[width=1.0\columnwidth,]{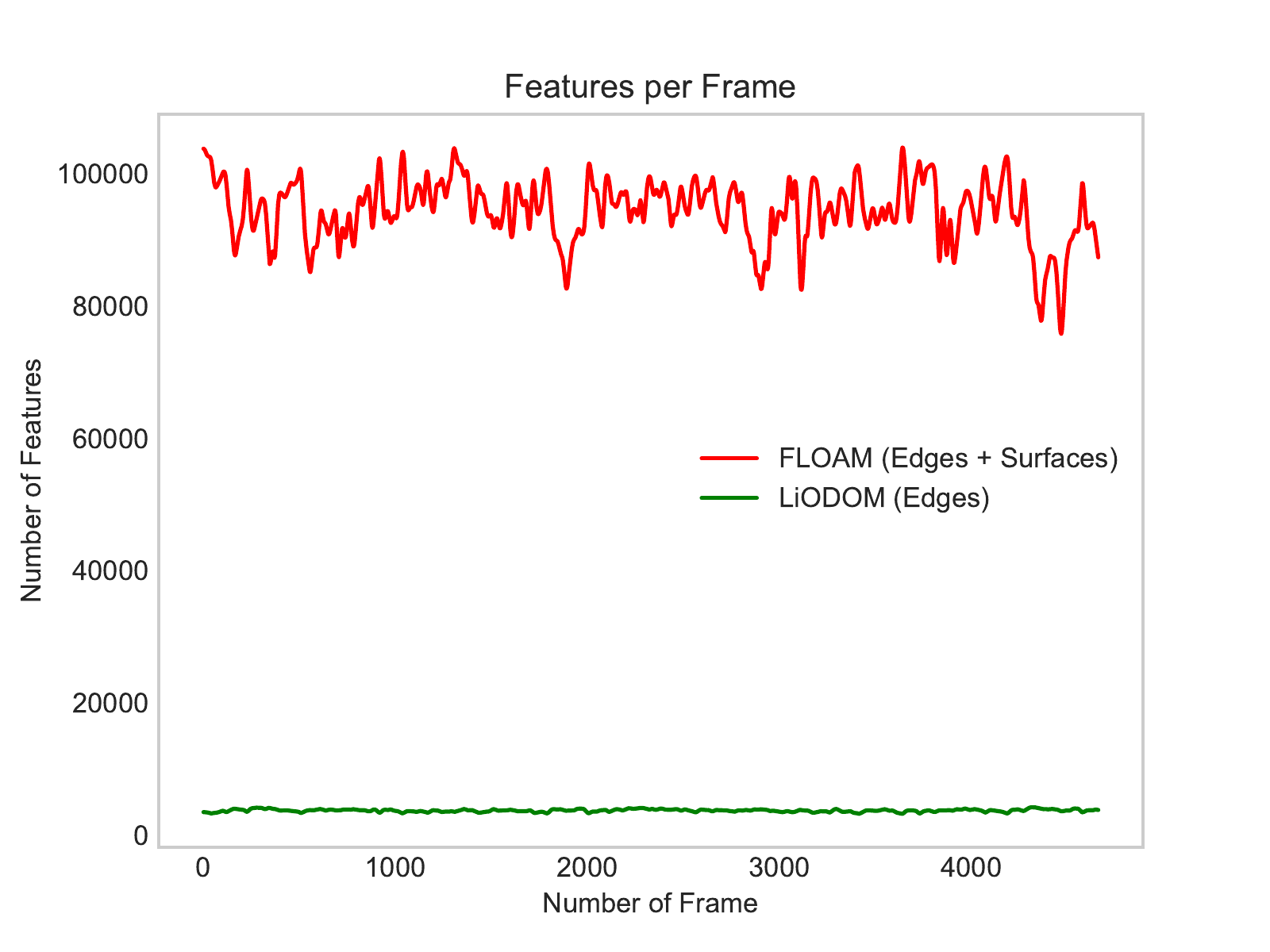}
\caption{Number of features per frame.}
\label{fig:feats_per_frame}
\end{figure}
Fig.~\ref{fig:feats_per_frame} reports on the number of features per frame that each solution has to handle. As can be observed, the differences between both approaches are more evident in this matter, since \appname{} only uses edges and, therefore, the required number of features is considerably less, with the corresponding reduction in memory requirements without noticeable differences in performance, slightly outperforming FLOAM for a number of datasets (as shown in Tables~\ref{tab:errors} and~\ref{tab:errorsATE}). 

\begin{table}[tb]
\small
\caption{\change{Absolute Trajectory Error (m) for the KITTI dataset with and without the weighting term.}}
\begin{center}
\begin{tabular}{c|cc}
& \textbf{With} & \textbf{Without} \\
\midrule
\textbf{K00} & 7.135  & 8.488  \\
\textbf{K02} & 9.754  & 11.613 \\
\textbf{K05} & 0.322  & 4.057  \\
\textbf{K06} & 0.956 & 1.024  \\
\textbf{K07} & 1.518  & 1.638  \\
\textbf{K08} & 4.592  & 5.780  \\
\textbf{K09} & 0.470  & 3.632  \\
\midrule
\textbf{Average} & 3.535 & 5.176\\
\end{tabular}
\end{center}
\label{tab:ATEweight}
\end{table}


\change{To finish, we consider the effect of the weighting term defined in Eq.~\ref{eq:weight}, which is related to the range returned by the LiDAR, as this results to be a key component of the pose optimization procedure of \appname. In this regard, Table~\ref{tab:ATEweight} reports on again the ATE for the KITTI sequences that contain loop closures, but with and without using the weighting factor. As can be observed, the weighting term improves the performance of \appname{} in all cases: all in all, the use of the weighting factor results into an average ATE of 3.535~m, in contrast to \appname{} without using this term, which leads to an average ATE of 5.176~m.}

\subsection{Mapping Performance}
\label{sec:mappingperf}

\begin{figure}[tb]
\centering
\includegraphics[width=1.0\columnwidth,]{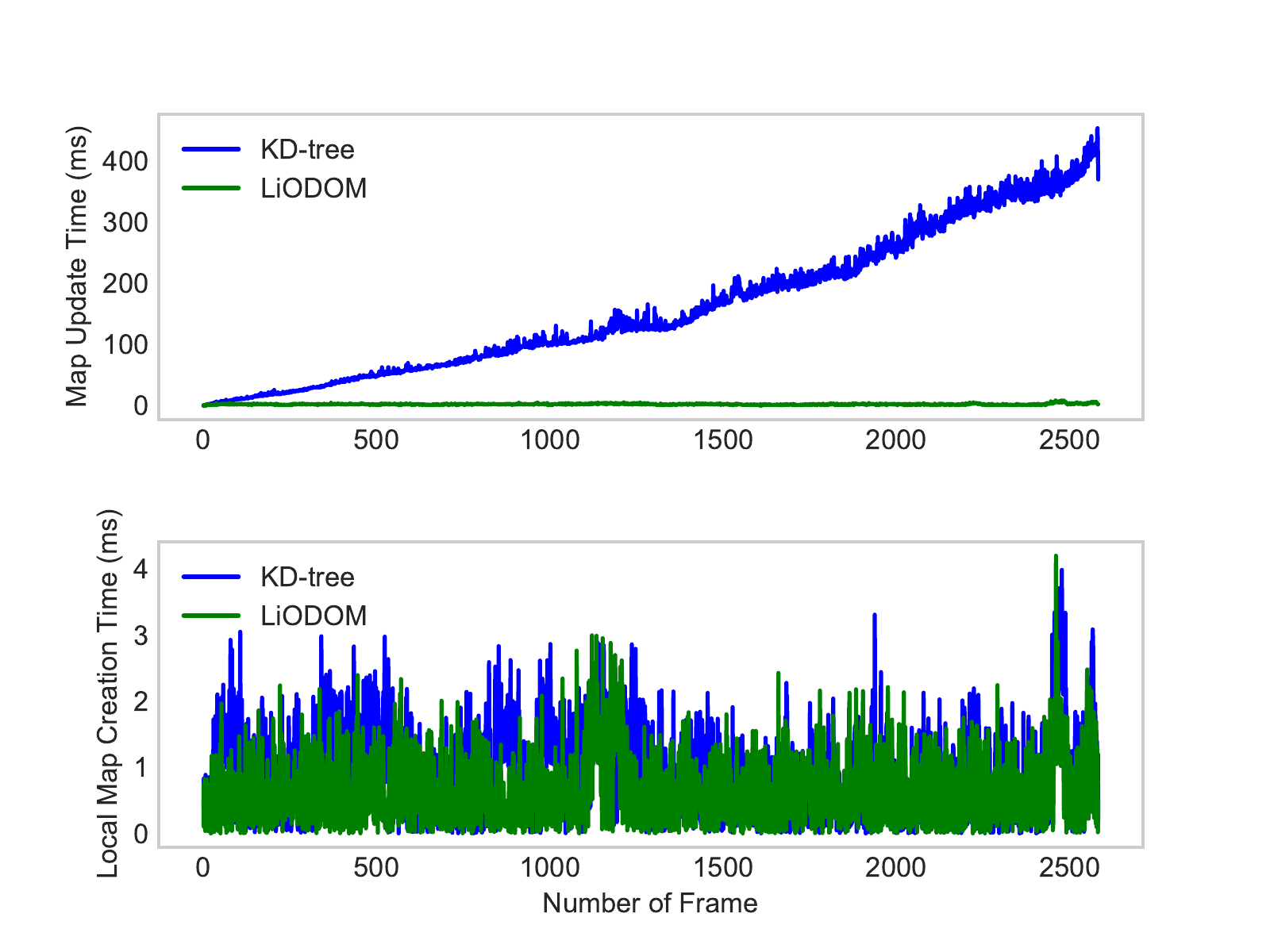}
\caption{Performance of the \appname{} mapping structure vs. a KD-tree.}
\label{fig:mappingtimes}
\end{figure}

In this section, we report on \change{several results} intended to assess the efficiency of the mapping approach adopted in \appname{}. In \change{a first} experiment, we measure the times required to update the global map and to build the local maps using our hashing-based data structure and a KD-tree. The K05 sequence was chosen in this case for computational reasons. The results are shown in Fig.~\ref{fig:mappingtimes}. As can be observed, the time required to update the global map by our approach remains roughly constant along the whole sequence. Contrarily, the running times for the KD-tree approach grow as more frames are processed, which can lead to an impractical operation. This behaviour can be attributed to the fact that, unlike our approach, the whole tree needs to be rebuilt on each update. The differences are less evident as for the times required to build the local maps, where both approaches are very fast, although our approach seems to perform slightly better.

\begin{table}[tb]
\small
\caption{\change{Entropy values for the hash table entry lengths for the KITTI sequences including loops.}}
\begin{center}
\begin{tabular}{c|cc}
& \textbf{Baseline} & \textbf{Eq.~\ref{eq:hashfun} (\appname{})} \\
\midrule
\textbf{K00} & 4.181  & 5.287  \\
\textbf{K02} & 4.409  & 5.645  \\
\textbf{K05} & 3.334  & 4.933  \\
\textbf{K06} & 3.766  & 4.464  \\
\textbf{K07} & 3.407  & 4.395  \\
\textbf{K08} & 3.457  & 5.321  \\
\textbf{K09} & 3.259  & 4.889  \\
\midrule
\textbf{Average} & 3.688 & 4.990\\
\end{tabular}
\end{center}
\label{tab:entropy}
\end{table}

\change{As a last experiment, in this section, we evaluate the quality of \appname{} hash function and the frequency of collisions when storing map points. In the ideal case, the hash function should be such that the number of elements associated to each bucket, i.e. the hash table entry lengths, follow a uniform distribution, what means that the hash function spreads reasonably well the elements between the table entries and therefore the number of collisions when accessing the map for updates is at a minimum. Given a distribution probability, its entropy $H$ can be used to check how close it is to the uniform distribution. For the case of the hash table, $H$ would be defined as:
    \begin{equation}
    H(m) = - \sum\limits_{k=1}^n p(k)\,\log(p(k))\,,
    \end{equation}
being $n$ the number of buckets in the hash table and $p(k)$ the probability that a new map point lies at the $k$-th table entry. $p(k)$ can thus be calculated as the number of data items associated to the entry divided by the total number of items stored in the table. The higher the value of the entropy $H$, the more uniformly are spread the points in the hash table and, therefore, the lower the probability of collision. Under this context, in this experiment, we compute the entropy $H$ of the global map generated by \appname{} after processing a full sequence. We compare the \appname{} hash function, i.e. Eq.~\ref{eq:hashfun}, with a baseline consisting on simply adding the hashes of the three cell coordinates. We employ again the sequences of the KITTI dataset including loops, since they are typically more prone to produce collisions given that they revisit previously seen places. The results are shown in Table~\ref{tab:entropy}. As can be observed, \appname{} in combination with the proposed hash function leads to higher entropy values than the alternative, simpler function in all cases.
}

\subsection{Experiments on-board an Aerial Platform}
\label{sec:mavexperiment}

\begin{figure}[tb]
    \begin{center}
        \begin{tabular}{cc}
        \includegraphics[width=0.58\columnwidth, height = 2.9cm, clip,trim=0cm 0cm 0cm 0cm]{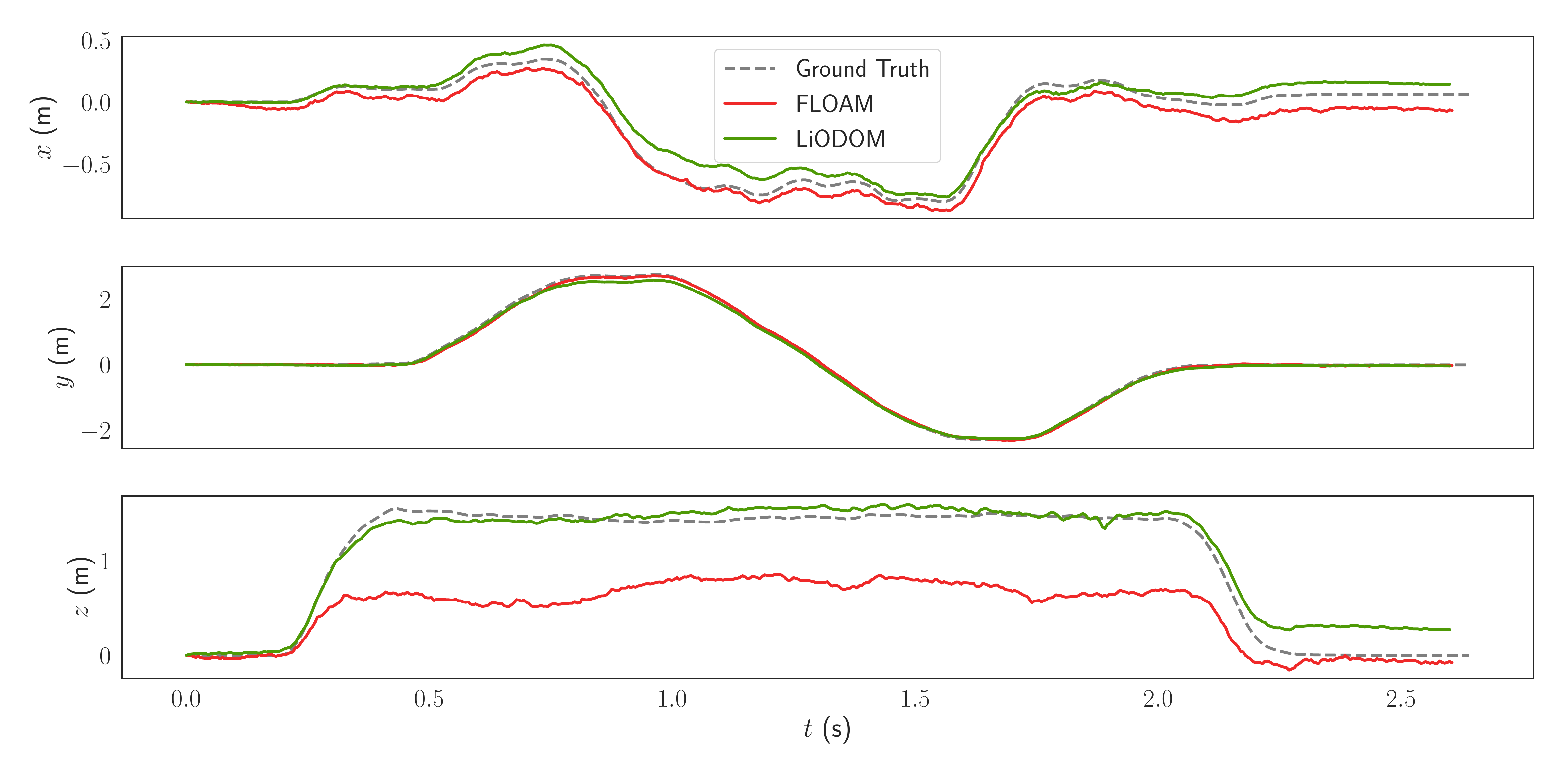}&
        \includegraphics[width=0.4\columnwidth,height = 2.9cm,clip,trim=18cm 0cm 2cm 3cm]{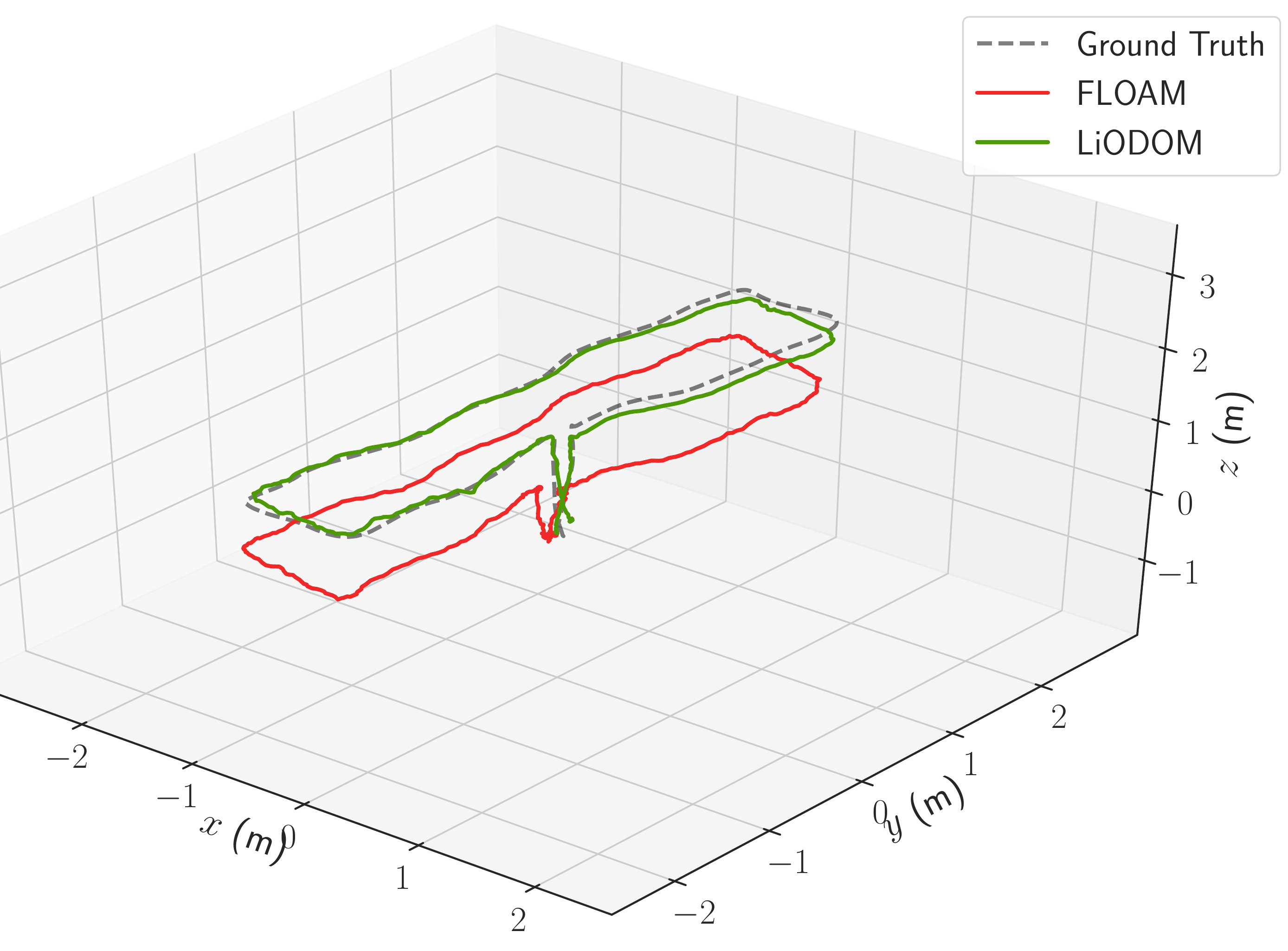}\\
       \includegraphics[width=0.58\columnwidth, height = 2.9cm, clip,trim=0cm 0cm 0cm 0cm]{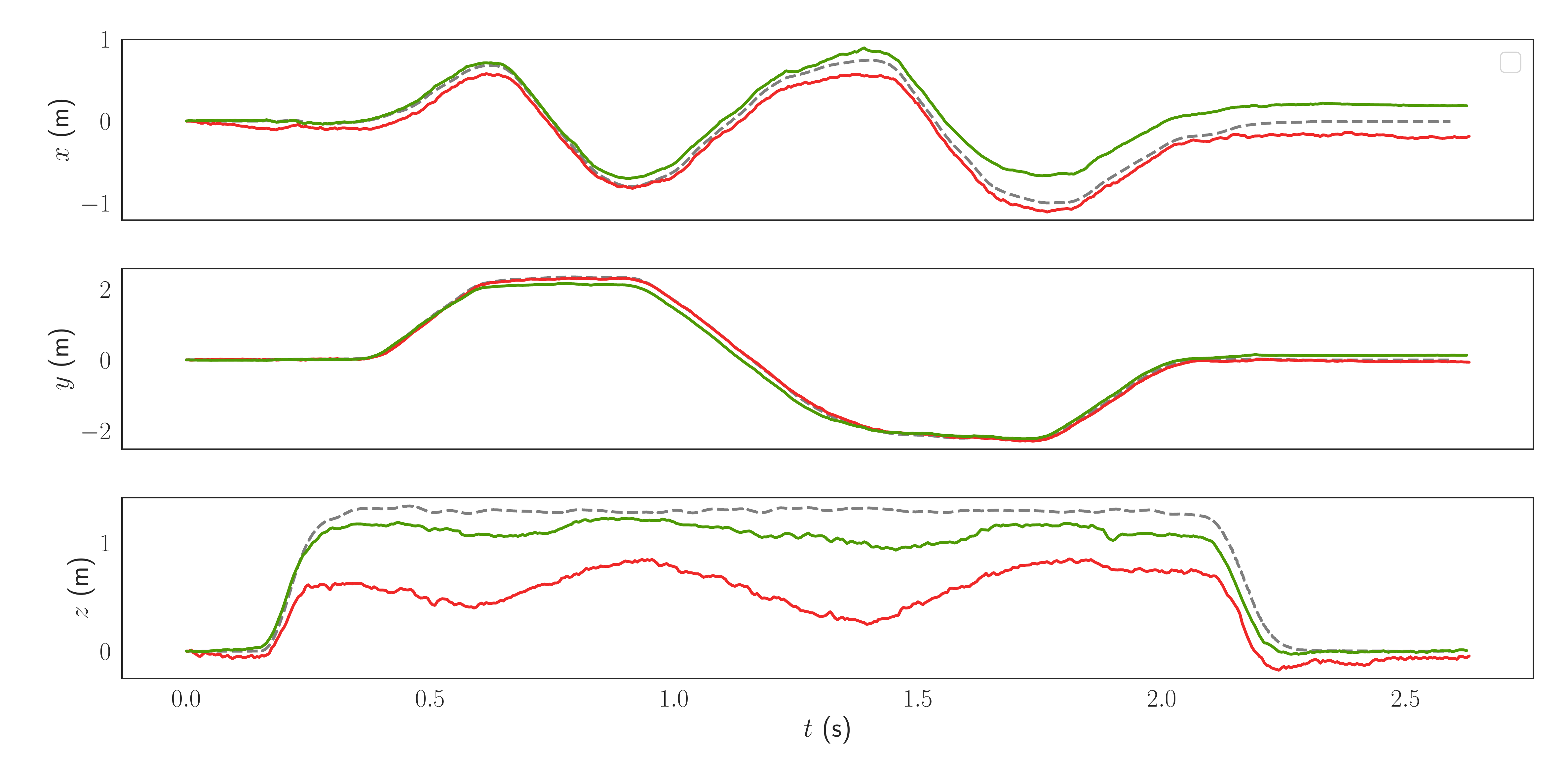}&
        \includegraphics[width=0.4\columnwidth,height = 2.9cm,clip,trim=18cm 0cm 2cm 3cm]{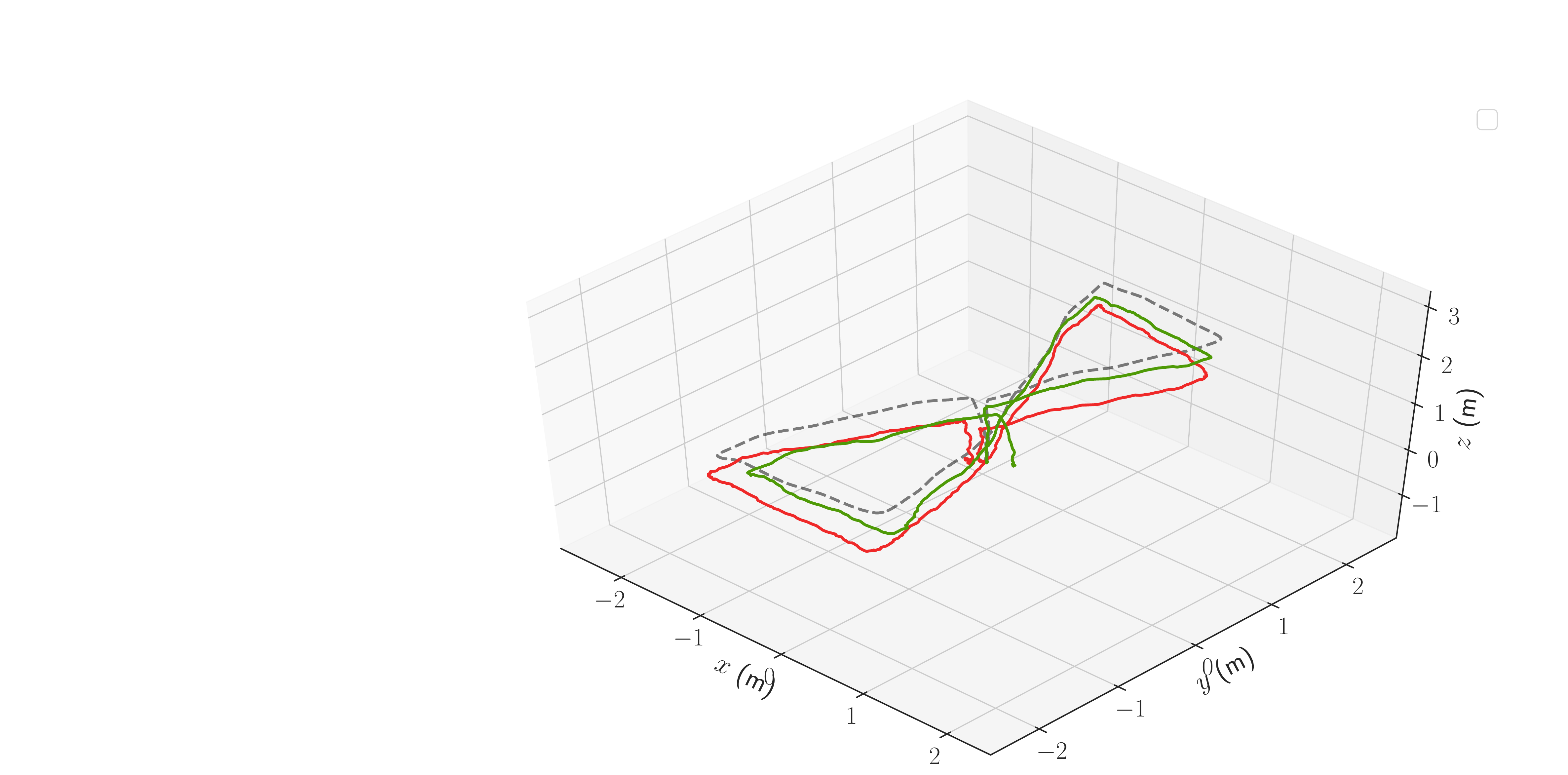}\\
        \includegraphics[width=0.58\columnwidth, height = 2.9cm, clip,trim=0cm 0cm 0cm 0cm]{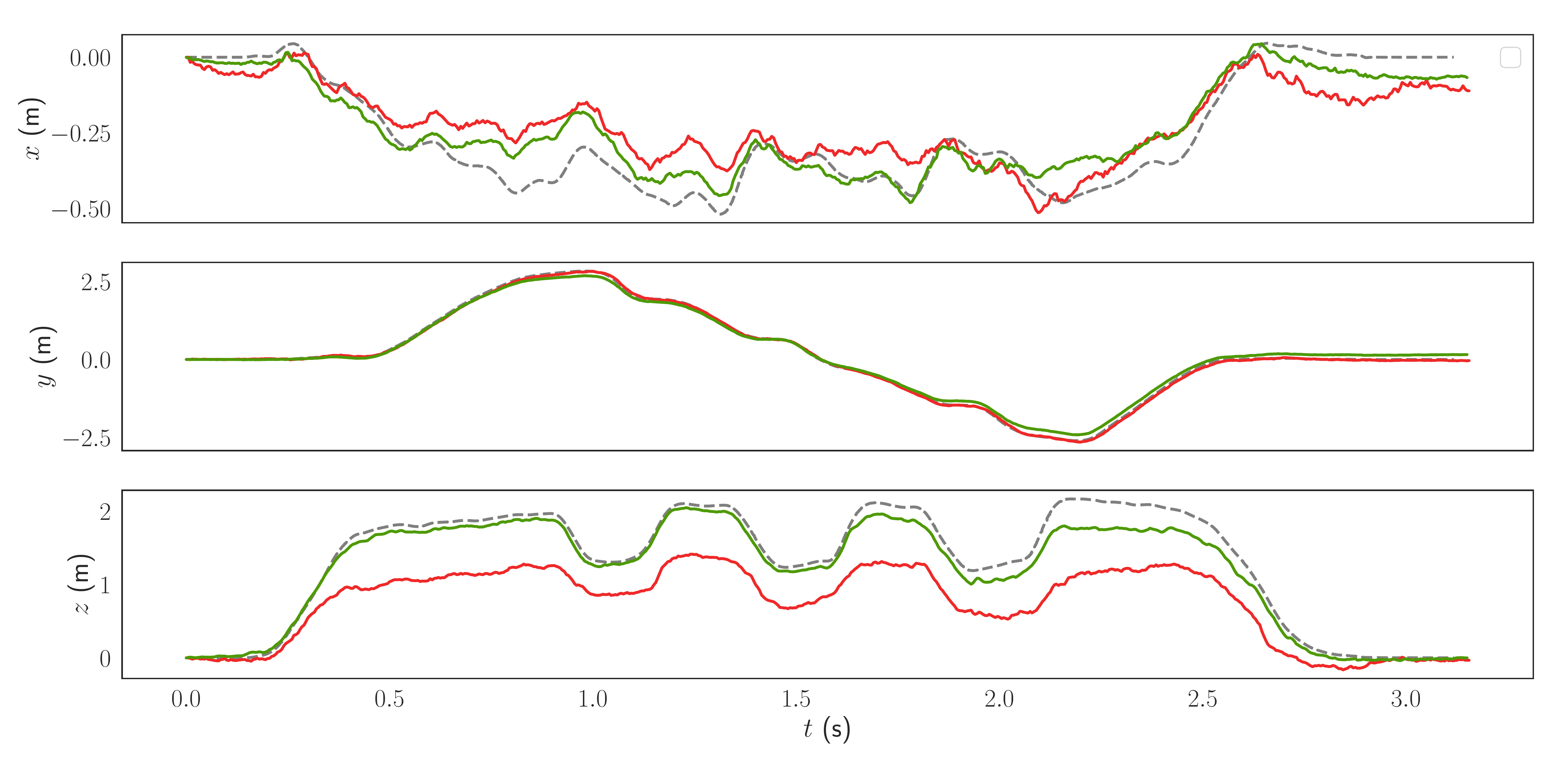}&
        \includegraphics[width=0.4\columnwidth,height = 2.9cm,clip,trim=20.9cm 0cm 2cm 3cm]{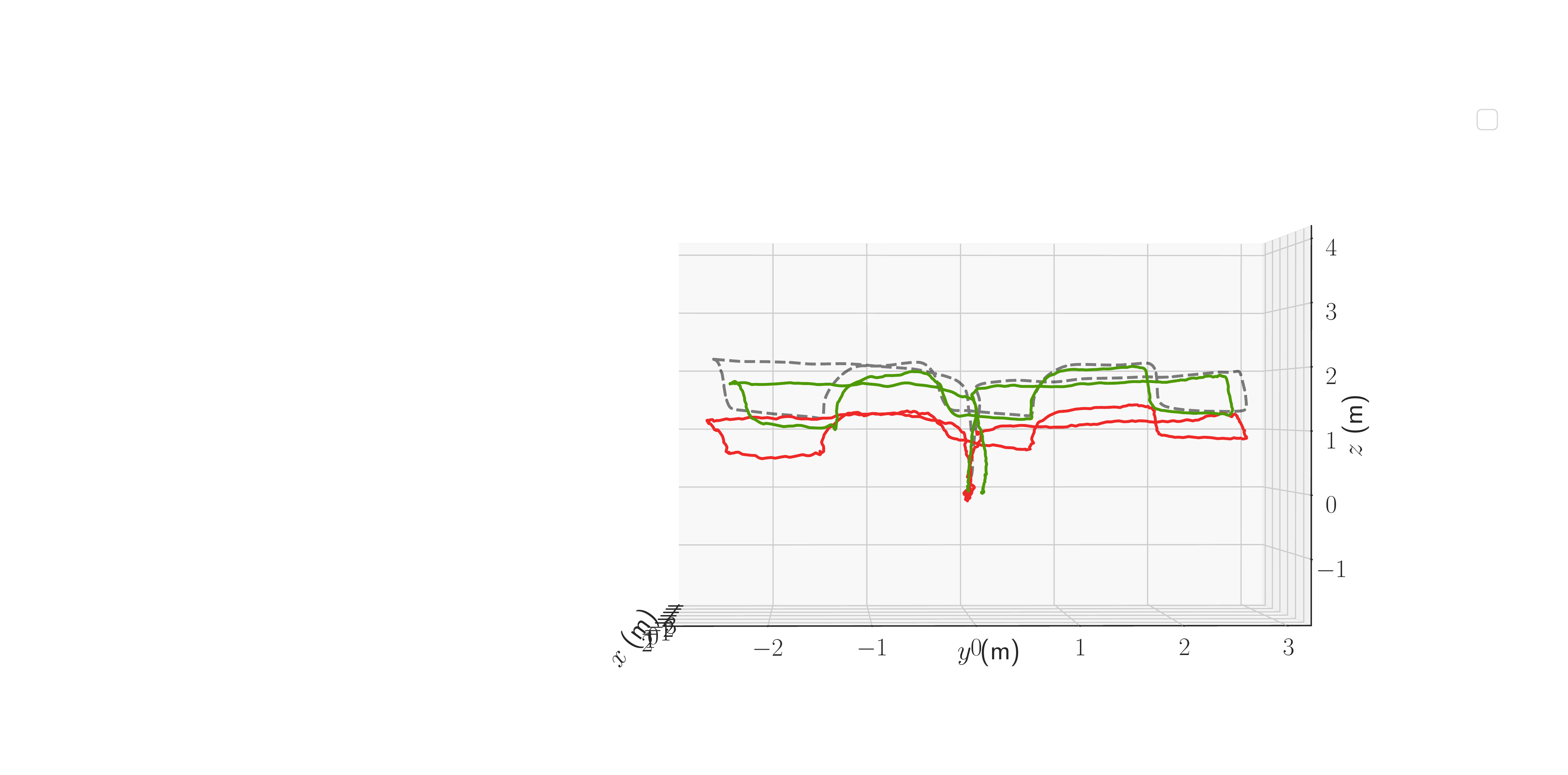}\\
        \includegraphics[width=0.58\columnwidth, height = 2.9cm, clip,trim=0cm 0cm 0cm 0cm]{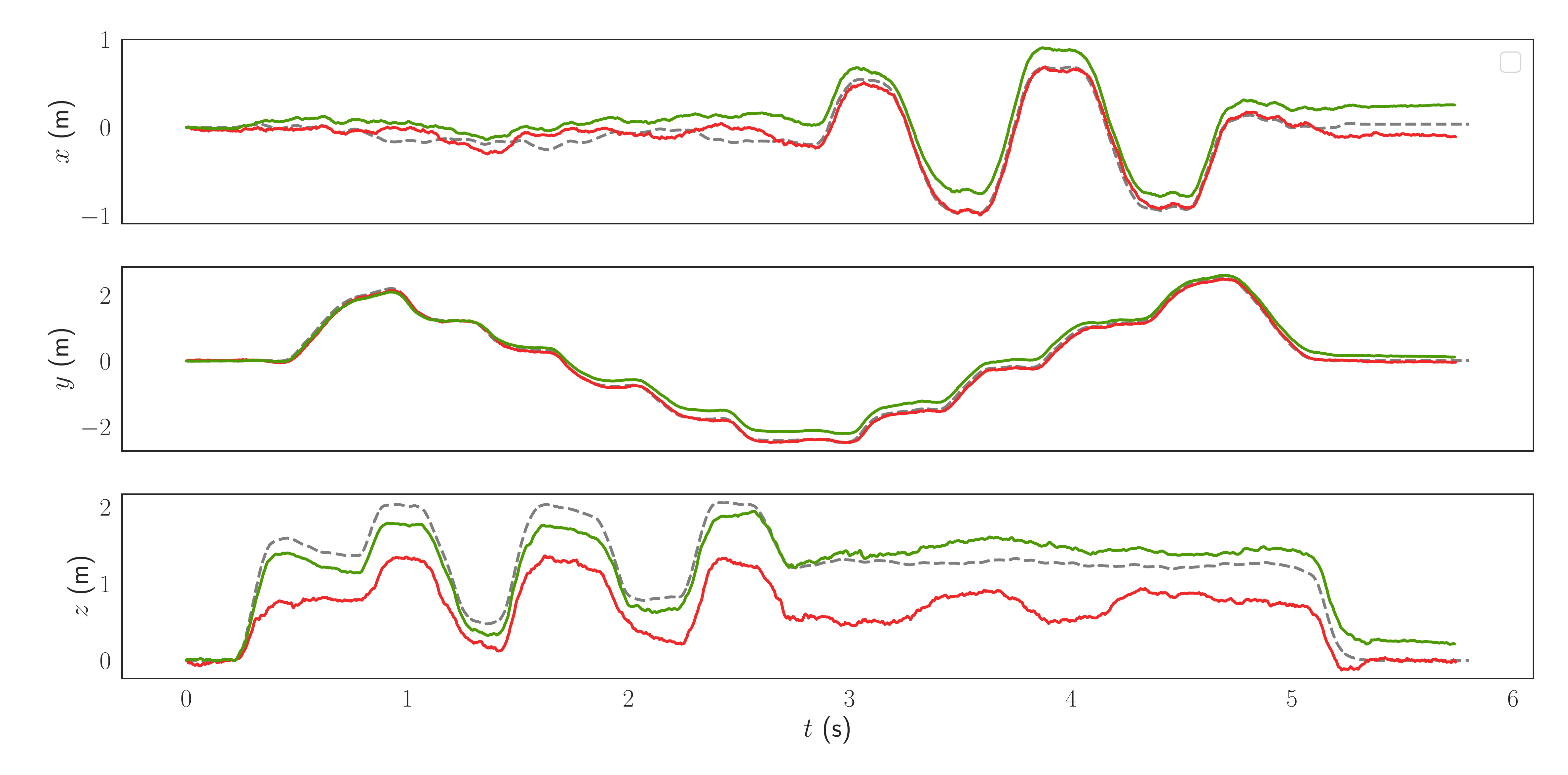}&
        \includegraphics[width=0.4\columnwidth,height = 2.9cm,clip,trim=18cm 0cm 2cm 3cm]{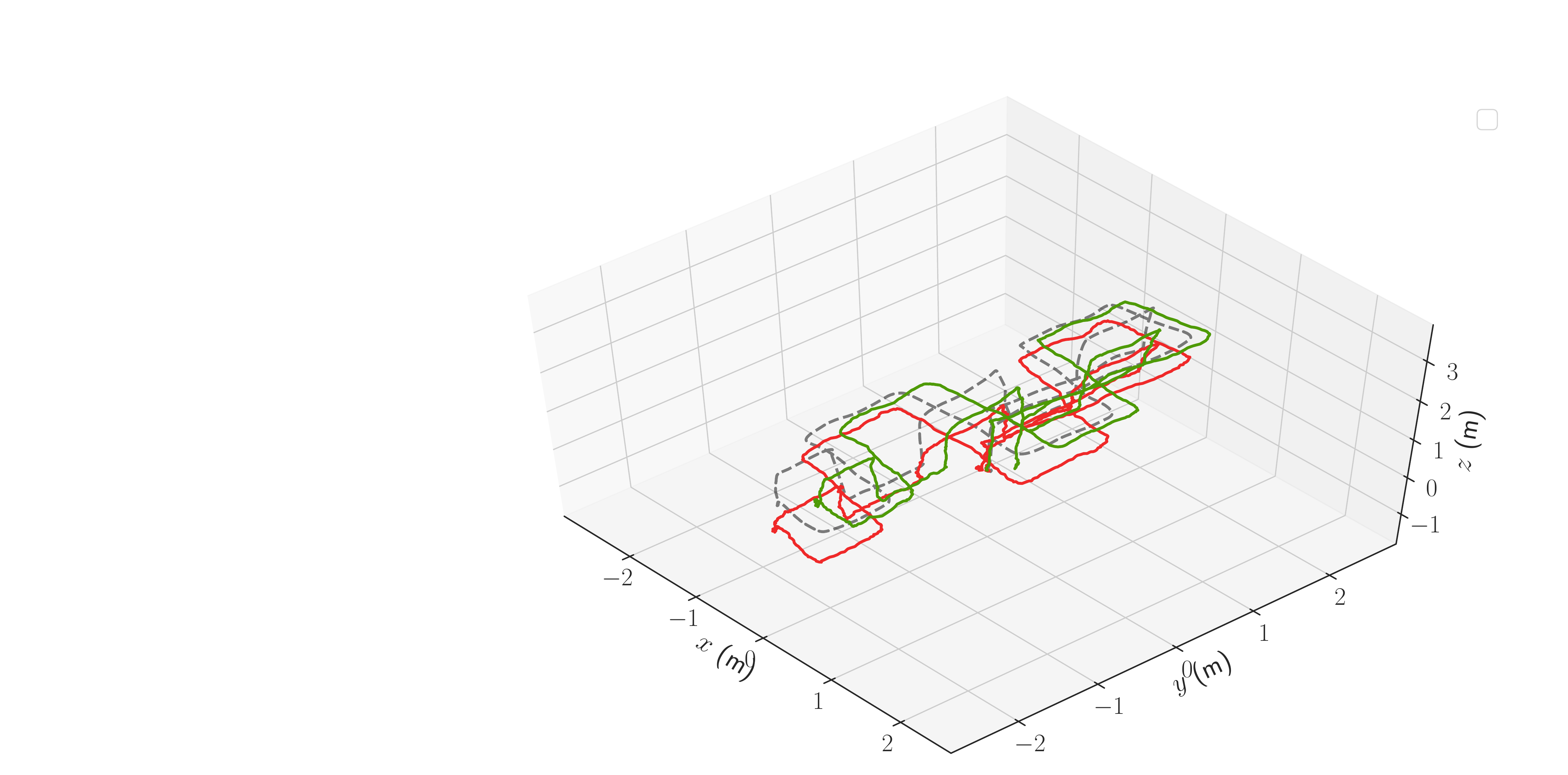}\\
        \includegraphics[width=0.58\columnwidth, height = 2.9cm, clip,trim=0cm 0cm 0cm 0cm]{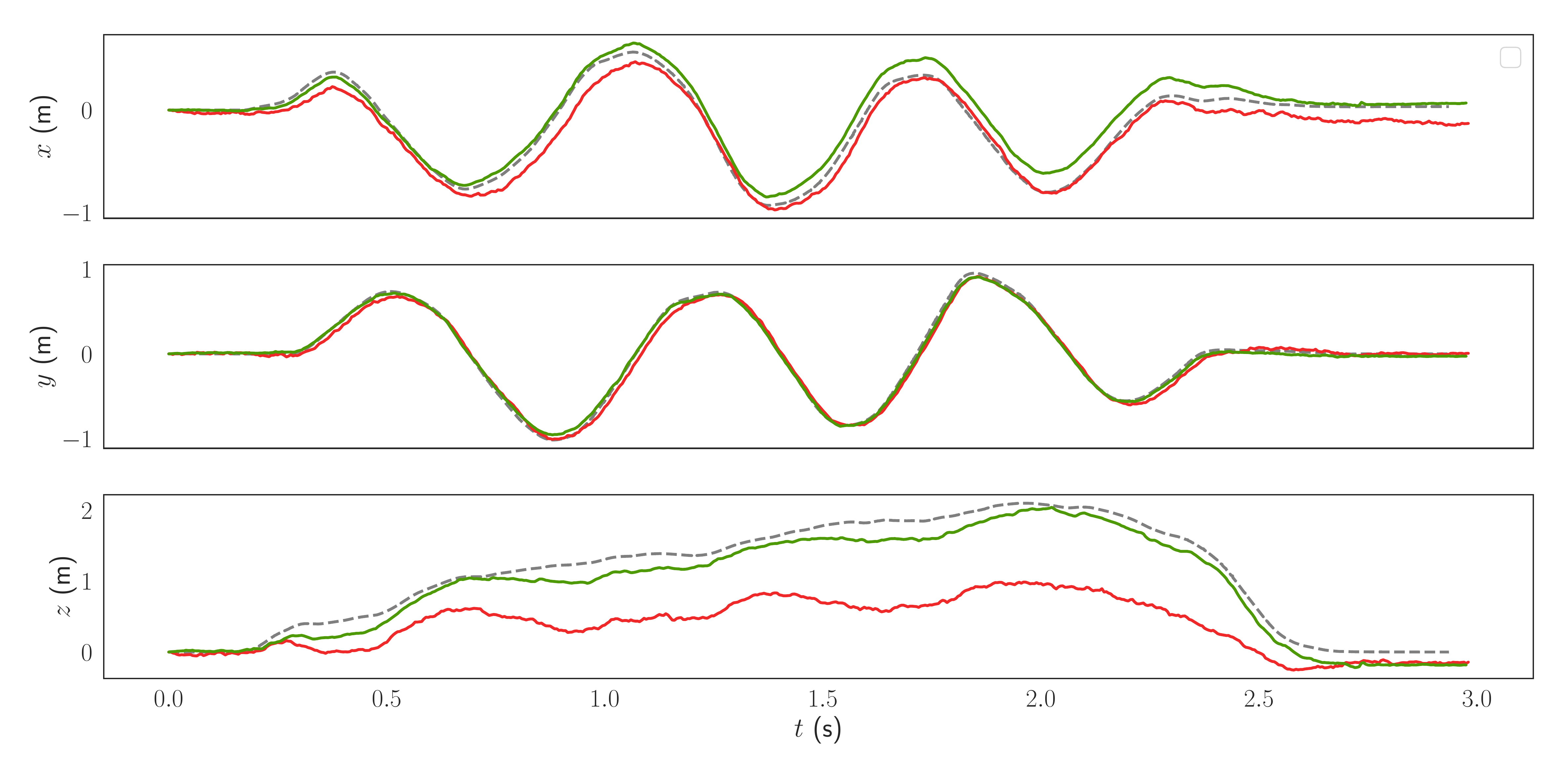}&
        \includegraphics[width=0.4\columnwidth,height = 2.9cm,clip,trim=18cm 0cm 2cm 3cm]{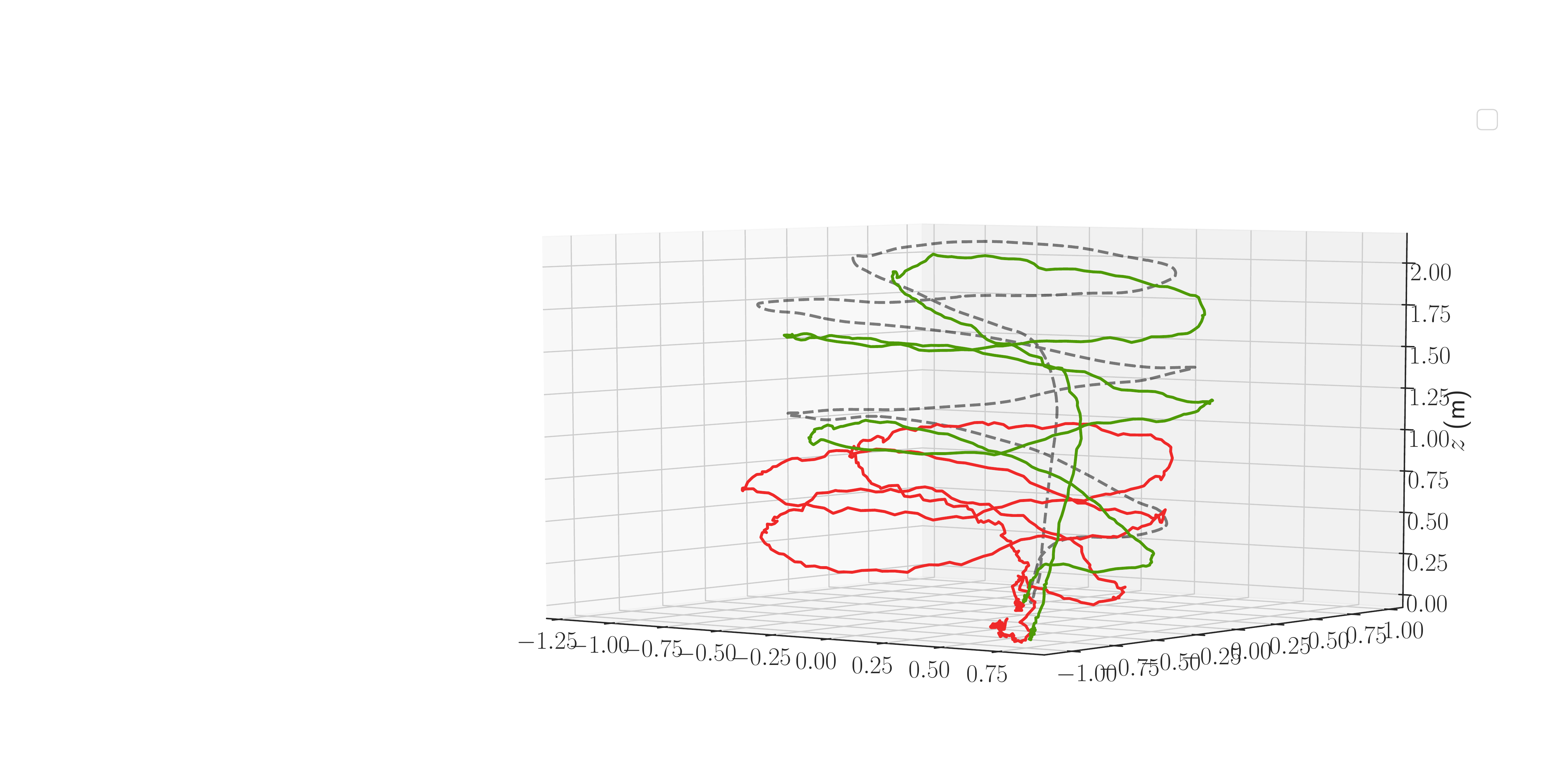}\\
        \includegraphics[width=0.58\columnwidth, height = 2.9cm, clip,trim=0cm 0cm 0cm 0cm]{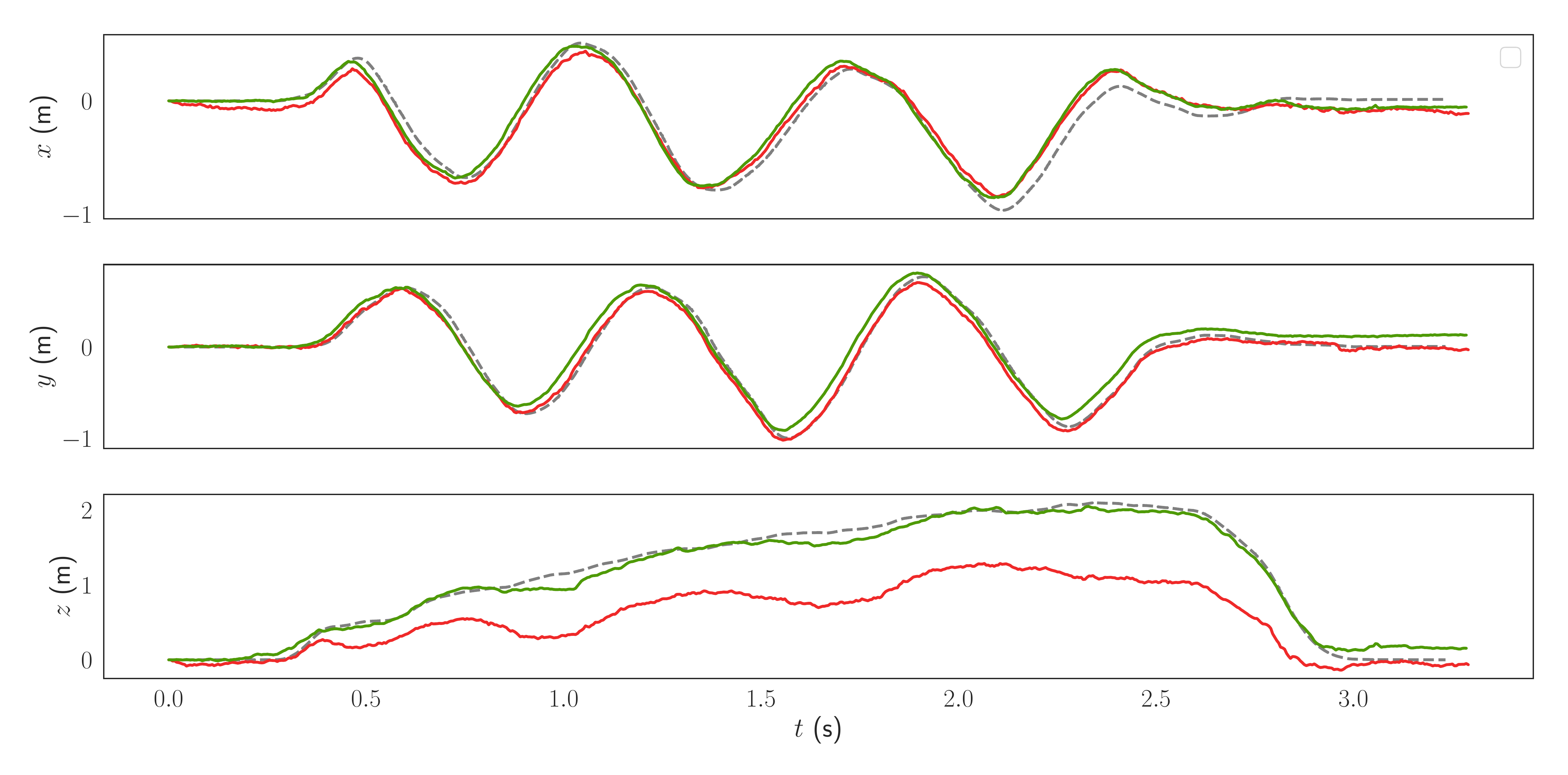}&
        \includegraphics[width=0.4\columnwidth,height = 2.9cm,clip,trim=18cm 0cm 2cm 3cm]{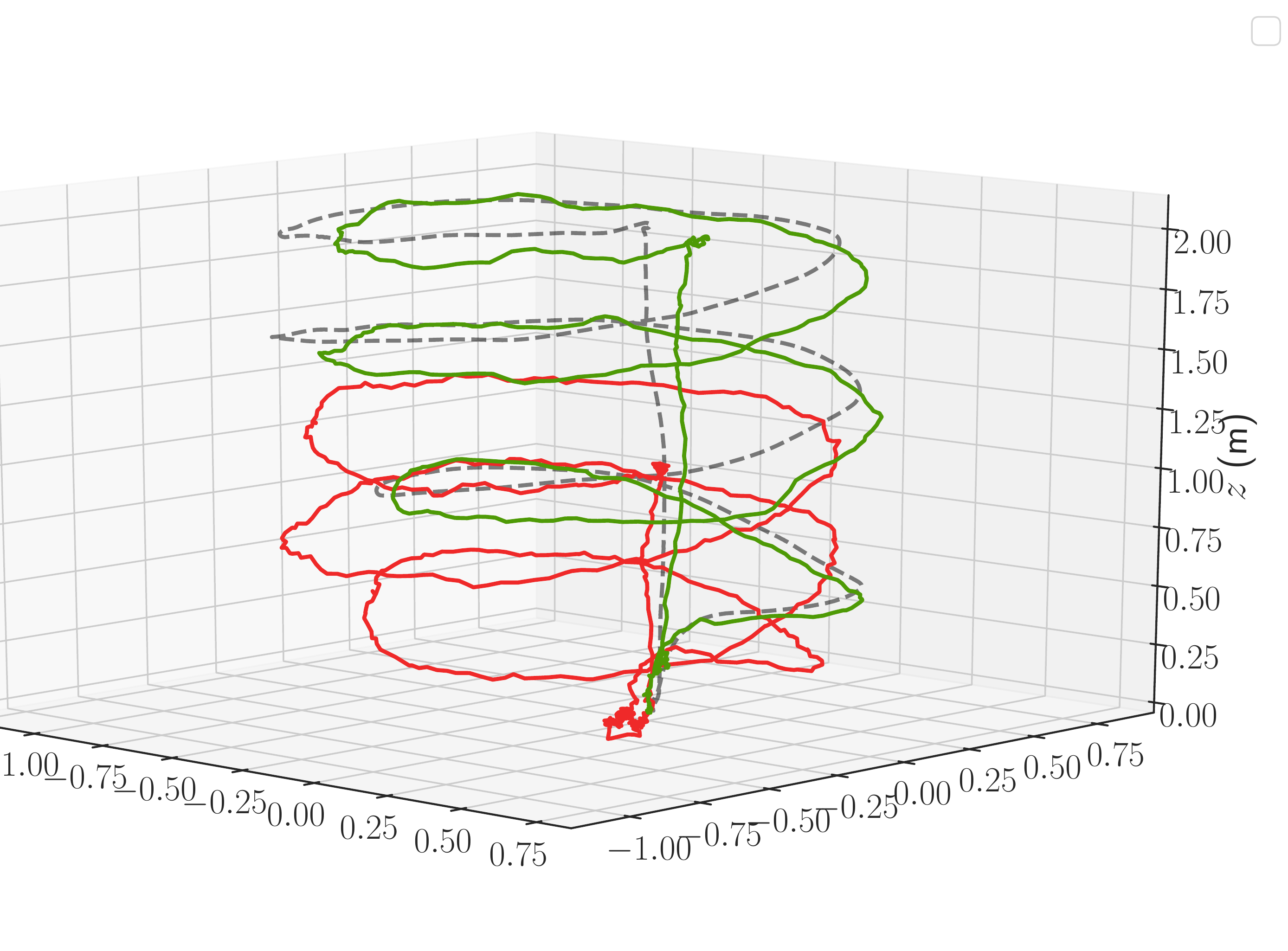}\\
        \end{tabular}
    \caption{\change{Position estimates and trajectories for the six experiments running \appname{} (green) and FLOAM (red) on-board an aerial platform.}}
    \label{fig:mav_pose_traj}
    \end{center}
\end{figure}

Finally, we also report on some experiments \change{involving} an aerial platform intended for visual inspection tasks~\cite{Bonnin-Pascual2021}. This platform has been recently fitted with an Ouster OS1-64 3D laser scanner that feeds \appname{}. The \change{performance evaluation} has been carried out inside the Aerial Robotics laboratory of the University of the Balearic Islands, which is equipped with an OptiTrack Motion Capture system (MOCAP) that supplies ground truth data during the tests. Table~\ref{tab:errorsMAV} shows the ATE \change{obtained using \appname{} and FLOAM} for six different experiments \change{mixing diverse motion in the three axes}. As can be noticed, the ATE values \change{from \appname{}} range from 16 to 31 cm, indicating that position estimates closely resemble the ground truth. \change{The ATE values from FLOAM are higher for all six experiments, ranging between 57 and 68 cm.} On the other side, Fig.~\ref{fig:mav_pose_traj} \change{compares, against the ground truth and along the respective trajectories, the position estimates of \appname{} and FLOAM} for all experiments. In this case, estimates for the X and Y axes mostly coincide with the ground truth, while, as also happens for other LiDAR-based odometry frameworks~\cite{Wang2021,Wang2020,Shan2018}, some drift can be appreciated in the Z-axis estimates. \change{Compared to the position estimates provided by FLOAM, the \appname{} estimates are closer to the ground truth, which is consistent with the lower ATE values shown in Table~\ref{tab:errorsMAV}}.



\begin{figure}[tb]
    \begin{center}
        \begin{tabular}{cc}
        \includegraphics[width=0.48\columnwidth,clip,trim=3cm 0cm 4.5cm 0cm]{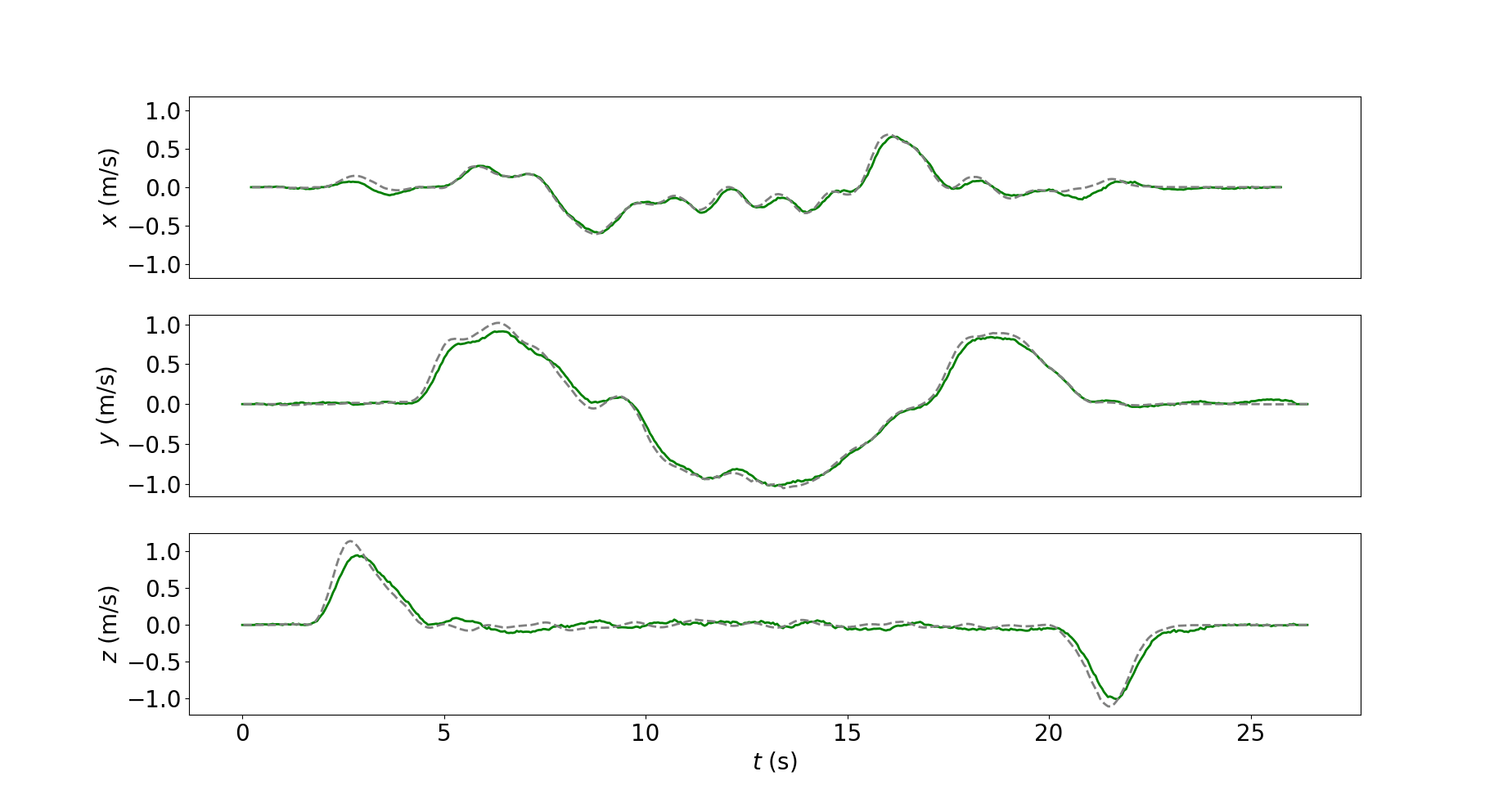}&
        \includegraphics[width=0.48\columnwidth,clip,trim=3cm 0cm 4.5cm 0cm]{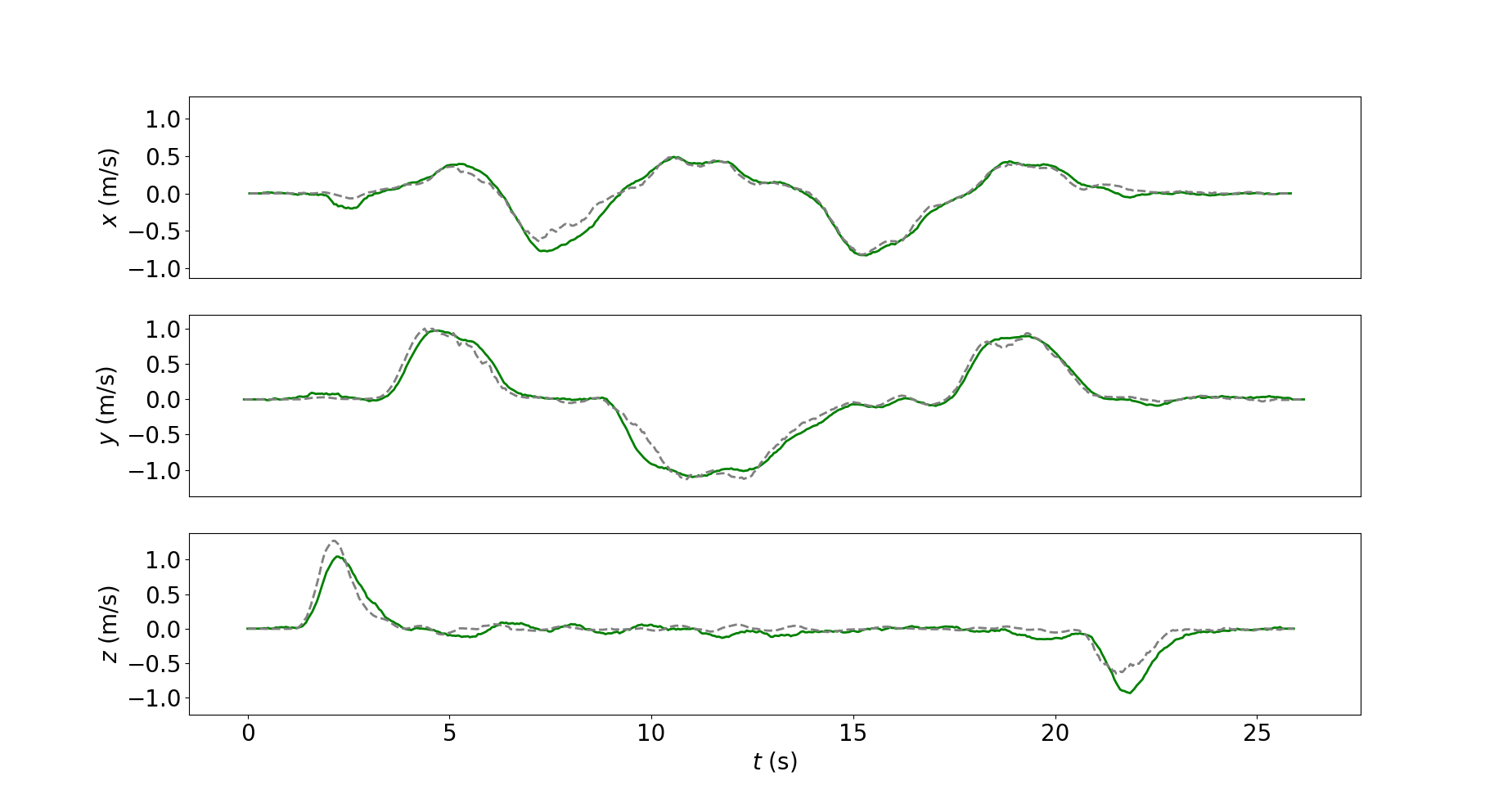}\\
        \includegraphics[width=0.48\columnwidth,clip,trim=3cm 0cm 4.5cm 0cm]{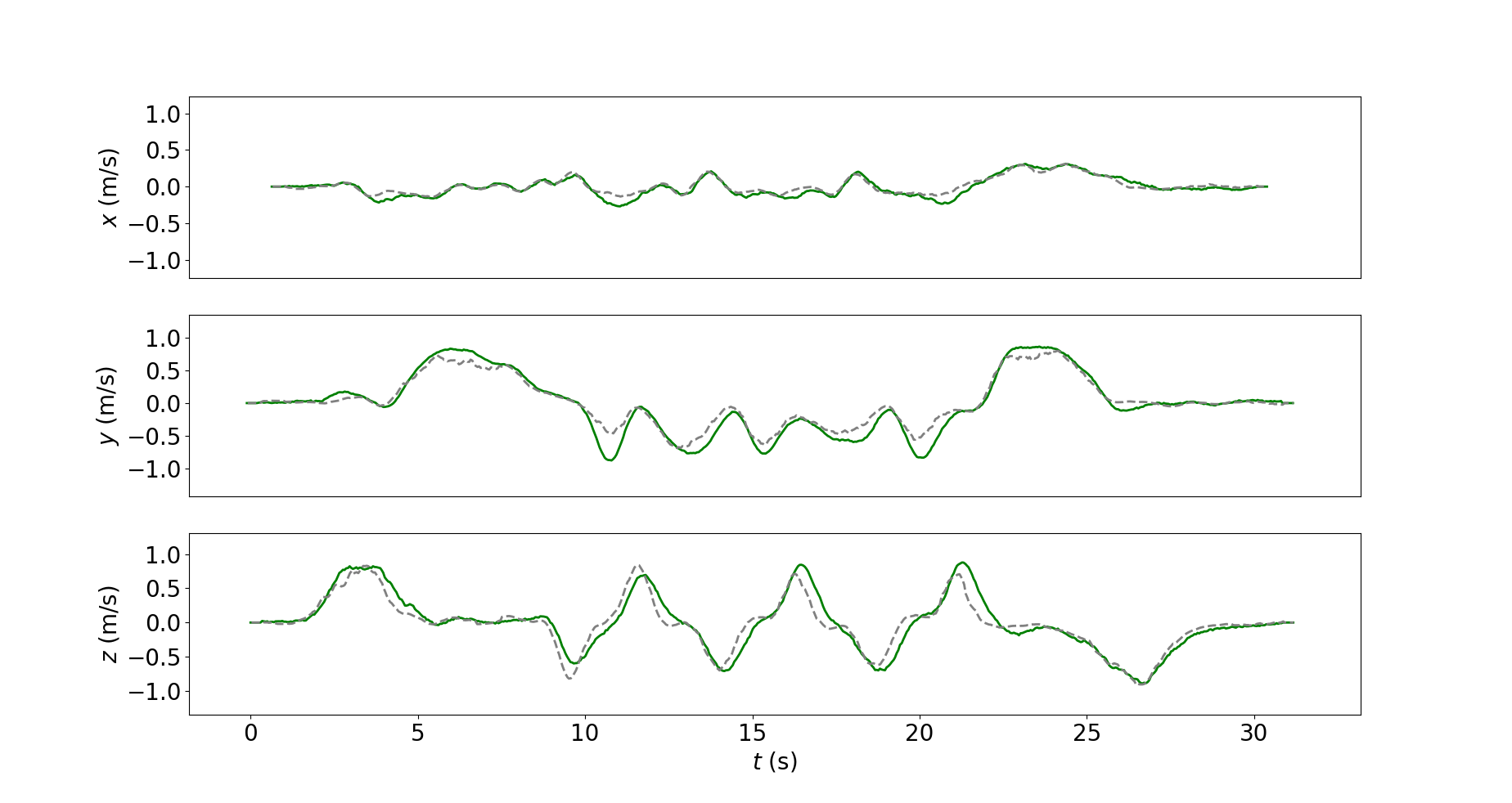}&
        \includegraphics[width=0.48\columnwidth,clip,trim=3cm 0cm 4.5cm 0cm]{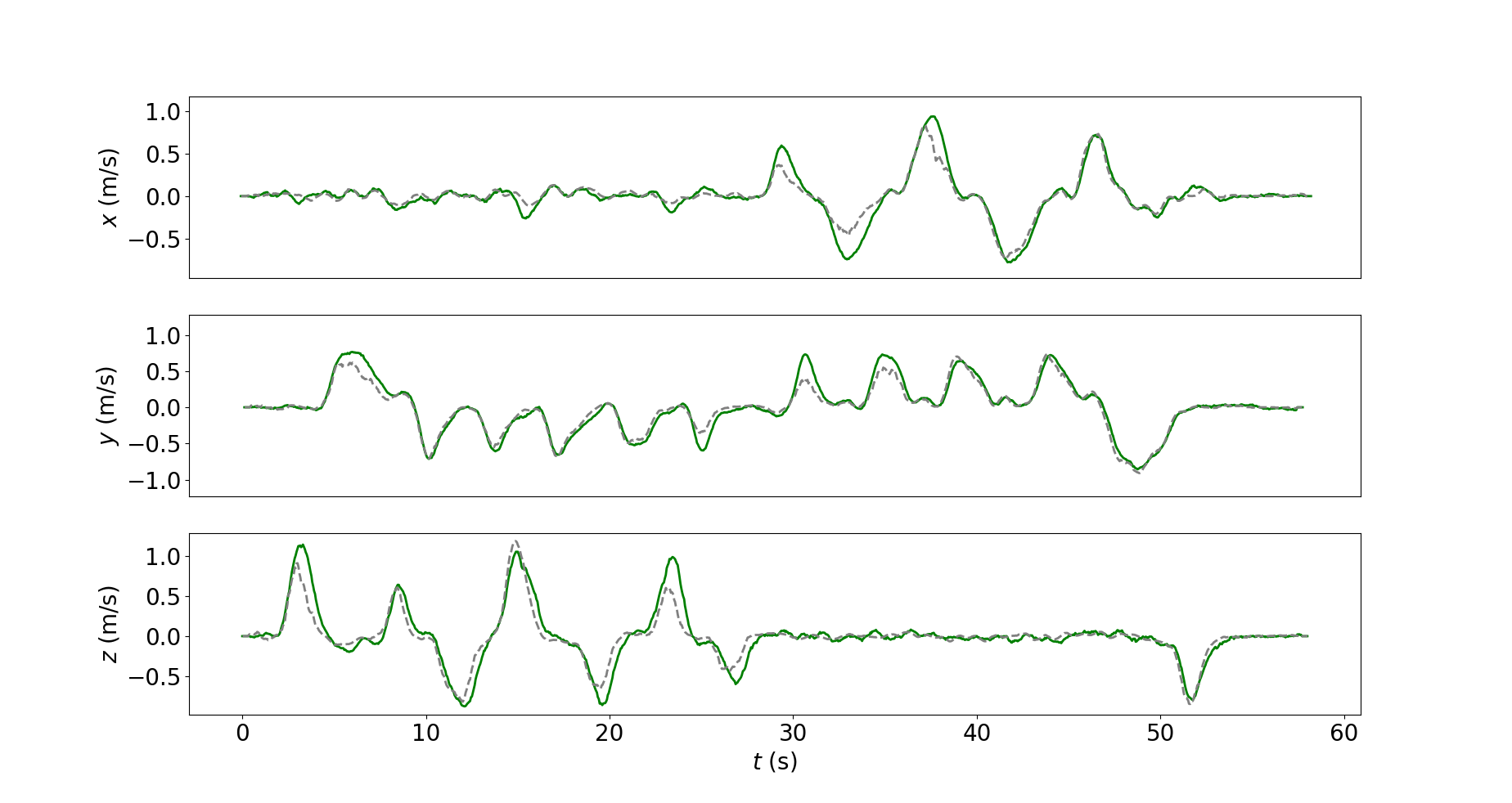}\\
        \includegraphics[width=0.48\columnwidth,clip,trim=3cm 0cm 4.5cm 0cm]{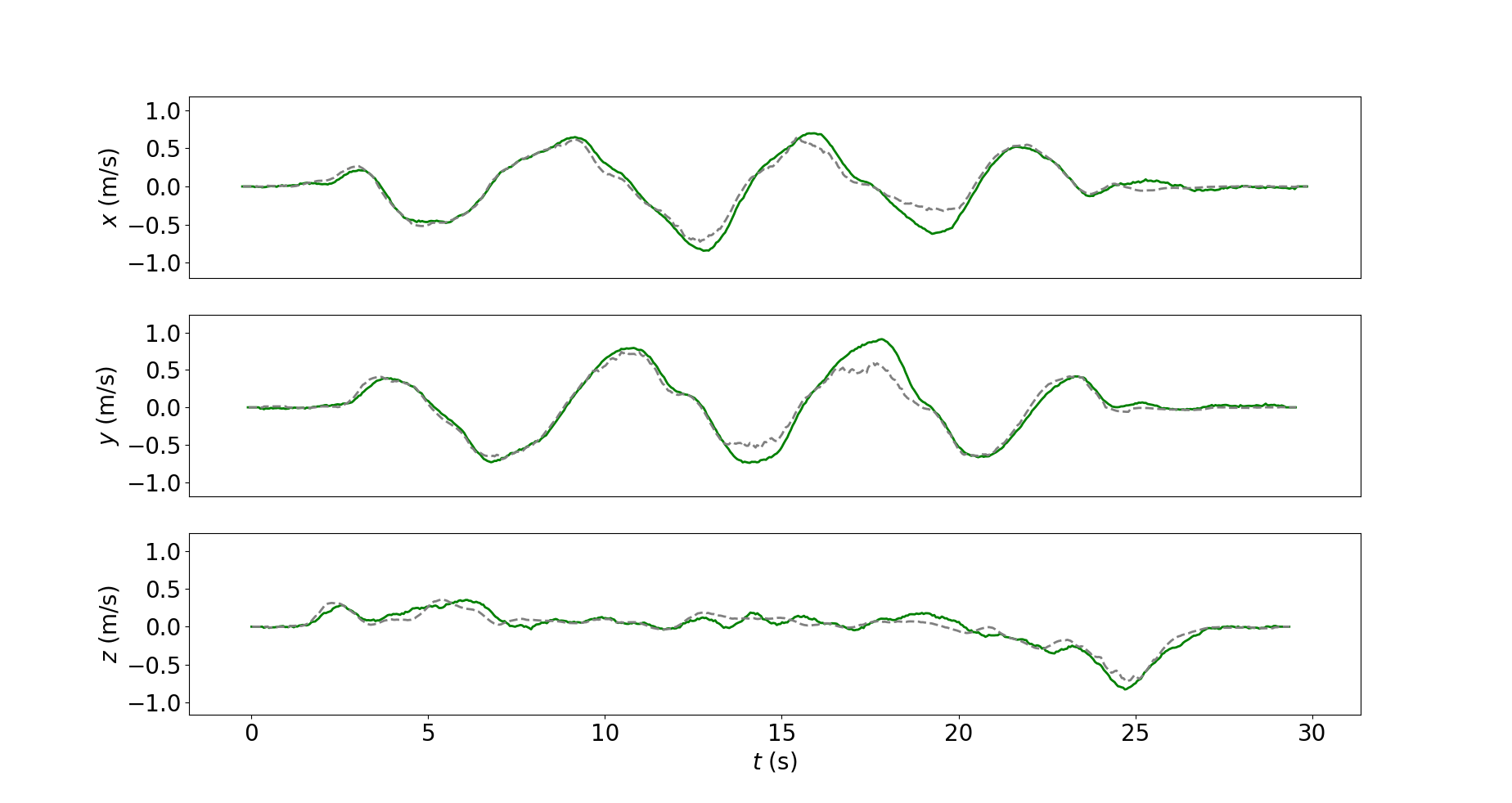}&
        \includegraphics[width=0.48\columnwidth,clip,trim=3cm 0cm 4.5cm 0cm]{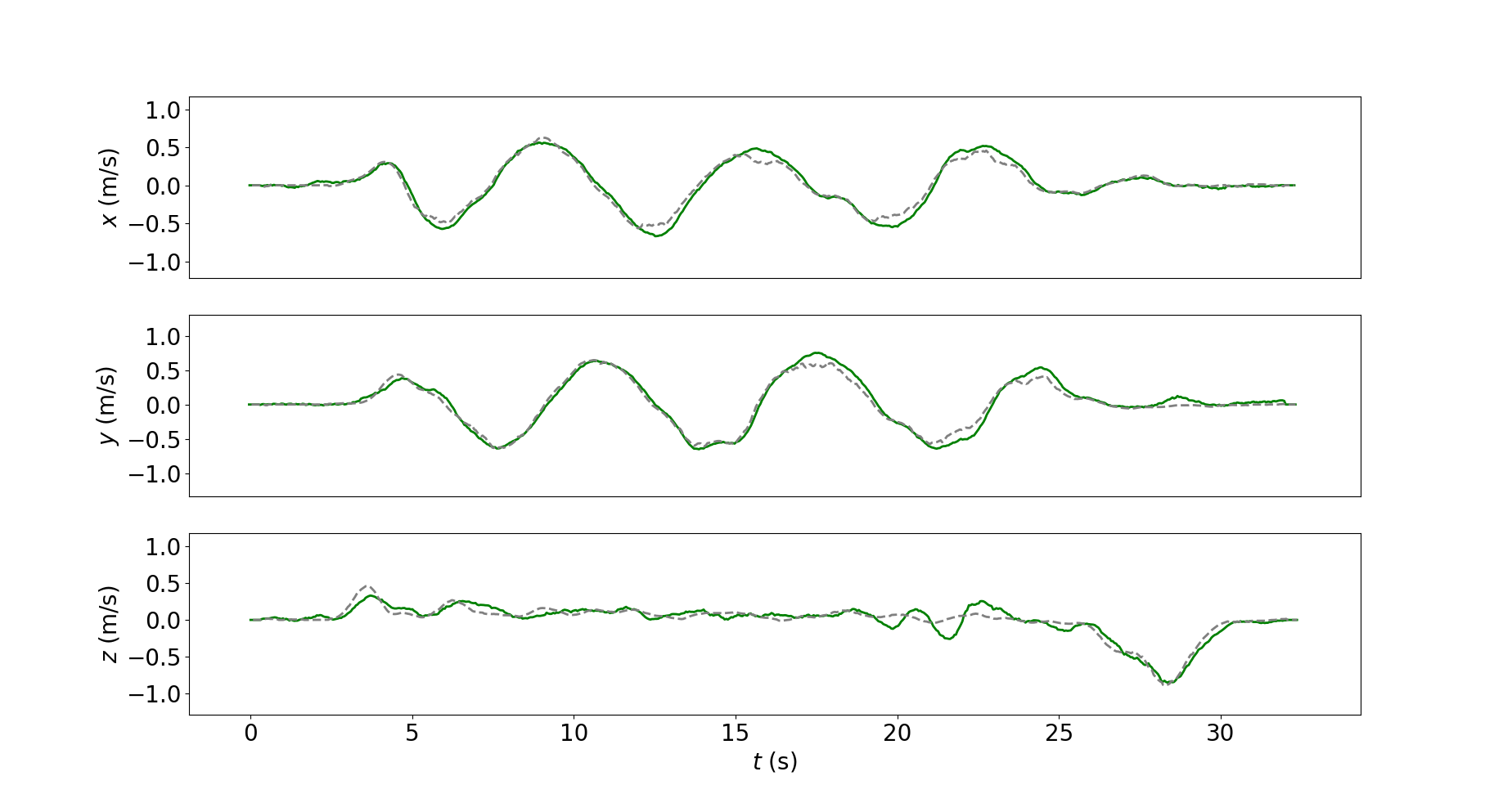}\\
        
        \end{tabular}
    \caption{\change{Velocity estimates for the experiments running \appname{} on-board an aerial platform. \appname{} estimates are shown in green, while the ground truth is shown as a grey dashed plot.}}
    \label{fig:vels_ex}
    \end{center}
\end{figure}

Within this robotic system, developed under the Supervised Autonomy paradigm, \appname{} is expected to supply not only pose estimates, but also velocity estimates, which constitute the basis for platform control in this case. \change{To show the performance of \appname{} in this regard,} Table~\ref{tab:errorsMAV} \change{additionally} reports on the velocity estimation results \change{against the ground truth} for the same six experiments as above, in the form of Root Mean Square Errors (RMSE) separately for each axis. The reported values indicate a very high accuracy in the estimation of X and Y velocities, and a slightly larger error for the Z axis. To finish, Fig.~\ref{fig:vels_ex} compares graphically the vehicle velocities estimated by \appname{} with the values provided by the MOCAP for the six experiments. As also observed for the position estimates, the X and Y velocity estimates coincide almost perfectly with the ground truth, while the Z-axis estimates present a slightly larger error.


\begin{table}[tb]
\small
\caption{\change{ATE (m) and velocity RMSE (m/s) for the experiments on-board a UAV}}
\begin{center}
\begin{tabular}{c||cc|ccc}
& \multicolumn{2}{c}{\textbf{ATE}} & \textbf{Velocity RMSE}\\
Experiment & \textbf{\appname{}} & \textbf{FLOAM} & \textbf{(x\,/\,y\,/\,z)}\\
\midrule
1 &  0.176776 & 0.657013 & 0.042 / 0.051 / 0.075 \\ 
2 & 0.276960 & 0.613781 & 0.071 / 0.079 / 0.101 \\ 
3 & 0.204852 & 0.637984 & 0.048 / 0.121 / 0.141 \\ 
4 & 0.314232 & 0.572278 & 0.088 / 0.085 / 0.129 \\ 
5 & 0.199806 & 0.816480 & 0.086 / 0.108 / 0.066 \\ 
6 & 0.164287 & 0.675760 & 0.063 / 0.068 / 0.074 \\ 
\bottomrule
\end{tabular}
\end{center}
\label{tab:errorsMAV}
\end{table}

\section{Conclusions and Future Work}
\label{sec:conclusions}

This paper proposes \appname{}, a novel LiDAR-only odometry and mapping approach. Our solution fundamentally consists of two parts working concurrently: (1) the odometry module, which is in charge of extracting a set of edges from the input sweep and estimating the current pose of the LiDAR; and (2) the mapping module, which builds and maintains a global map of the environment, and also generates a local map employed for pose estimation. Pose estimation is conceived as an optimization problem which involves a set of weighted point-to-line constraints between the current sweep and a local map. We have also described a data structure based on a hashing scheme which allows us to rapidly get access to any part of the map and manage it in an efficient way. Furthermore, this structure is also employed to obtain an adaptive local map, used to facilitate data association. Our experiments show that \appname{} compares favourably against other state-of-the-art approaches, and that it can be used for both position and velocity estimation.

Despite its good performance, \appname{} is an odometer and unavoidably drifts. Therefore, we will consider extending the ideas proposed in this paper to develop a complete SLAM / 3D reconstruction system, incorporating other motion estimation sensors into a fusion scheme for enhanced performance.

\appendix

\section*{Acknowledgments}
This work is partially supported by EU-H2020 projects BUGWRIGHT2 (GA 871260) and by project PGC2018-095709-B-C21 (funded by MCIU/AEI/ 10.13039/501100011033 and FEDER ``Una manera de hacer Europa"). This publication reflects only the authors views and the European Union is not liable for any use that may be made of the information contained therein.

\bibliographystyle{elsarticle-num} 
\bibliography{ms}

\end{document}